\documentclass[10pt,twocolumn,letterpaper]{article}

\usepackage[pagenumbers]{cvpr} 

\usepackage{graphicx}
\usepackage{amsmath}
\usepackage{amssymb}
\usepackage{booktabs}
\usepackage{caption}
\usepackage{enumitem}
\usepackage{multirow, makecell}
\usepackage{url}
\usepackage[misc]{ifsym}

\usepackage[accsupp]{axessibility}  

\usepackage[pagebackref,breaklinks,colorlinks]{hyperref}

\usepackage[capitalize]{cleveref}
\crefname{section}{Sec.}{Secs.}
\Crefname{section}{Section}{Sections}
\Crefname{table}{Table}{Tables}
\crefname{table}{Tab.}{Tabs.}

\newcommand{\ys}{\textcolor{black}} 

\def\cvprPaperID{647} 
\def\confName{CVPR}
\def\confYear{2022}

\begin{document}

\title{Pastiche Master: Exemplar-Based High-Resolution Portrait Style Transfer\vspace{-2mm}}

\author{Shuai Yang \hspace{12pt} Liming Jiang  \hspace{12pt}  Ziwei Liu  \hspace{12pt} Chen Change Loy$^{~\textrm{\Letter}}$\\
S-Lab, Nanyang Technological University\\
{\tt\small \{shuai.yang, liming002,  ziwei.liu, ccloy\}@ntu.edu.sg}
}

\twocolumn[{%
\renewcommand\twocolumn[1][]{#1}%
\maketitle
\vspace{-3em}
\begin{center}
\centering
\includegraphics[width=0.96\linewidth]{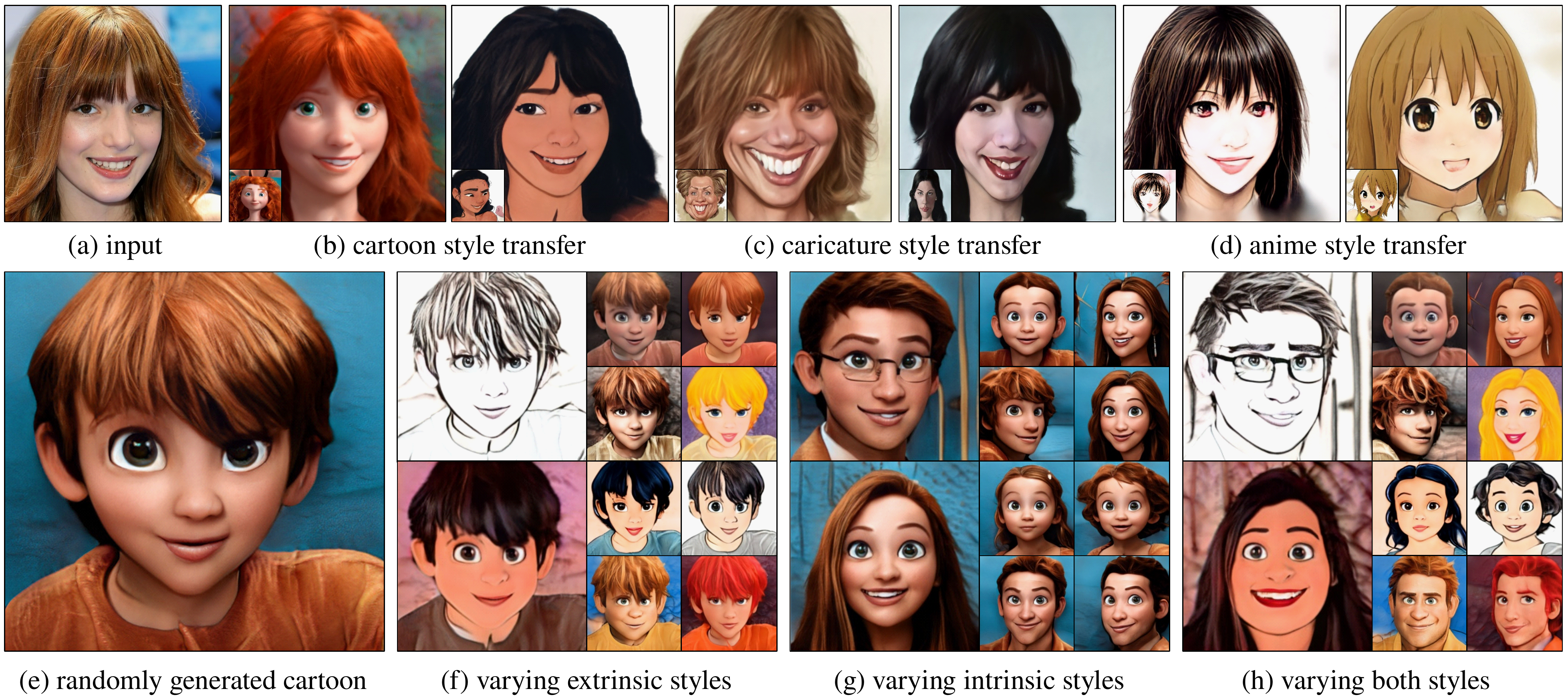}\vspace{-2mm}
\captionof{figure}{We propose a novel DualStyleGAN for exemplar-based high-resolution (1024$\times$1024) portrait style transfer. The artistic portraits of (b)-(d) generated from a real face (a) successfully imitate the color and structural styles of the examples seen in their respective lower-left corners.
DualStyleGAN features dual style paths: an intrinsic style path and an extrinsic style path for flexible control over the content and style, respectively. (e) A cartoon face generated from arbitrary intrinsic and extrinsic style codes. Samples generated by (f) varying extrinsic styles with fixed intrinsic styles, (g) varying intrinsic styles with fixed extrinsic styles, and (h) varying both styles.}\vspace{-1mm}
\label{fig:teaser}
\end{center}%
}]


\begin{abstract}\vspace{-1mm}
  Recent studies on StyleGAN show high performance on artistic portrait generation by transfer learning with limited data.
  In this paper, we explore more challenging exemplar-based high-resolution portrait style transfer by introducing a novel \textbf{DualStyleGAN} with flexible control of dual styles of the original face domain and the extended artistic portrait domain. Different from StyleGAN, DualStyleGAN provides a natural way of style transfer by characterizing the content and style of a portrait with an \textbf{intrinsic style path} and a new \textbf{extrinsic style path}, respectively. The delicately designed extrinsic style path enables our model to modulate both the color and complex structural styles hierarchically to precisely pastiche the style example. Furthermore, a novel progressive \ys{fine-tuning} scheme is introduced to smoothly transform the generative space of the model to the target domain, even with the above modifications on the network architecture. Experiments demonstrate the superiority of DualStyleGAN over state-of-the-art methods in high-quality portrait style transfer and flexible style control. \ys{Code is available at} \url{https://github.com/williamyang1991/DualStyleGAN}.\vspace{-1mm}
\end{abstract}

\section{Introduction}\vspace{-1.5mm}
\label{sec:intro}

Artistic portraits are popular in our daily lives and especially in industries related to comics, animations, posters, and advertising.
In this paper, we focus on exemplar-based portrait style transfer, a core problem that aims to transfer the style of an exemplar artistic portrait onto a target face.
Its potential application is appealing in that it allows any novice to easily transform their photograph into a stunning pastiche based on the style of their favourite artworks, which otherwise would have required highly professional skills for manual creation.

Automatic portrait style transfer based on image style transfer~\cite{Li2016Combining,selim2016painting,liao2017visual} and image-to-image translation~\cite{kim2019u,li2021anigan,chong2021gans} has been extensively studied. 
Recently, StyleGAN~\cite{karras2019style,karras2020analyzing}, the state-of-the-art face generator, has been very promising for high-resolution artistic portrait generation via transfer learning~\cite{pinkney2020resolution}.
Specifically, StyleGAN can be effectively fine-tuned, usually only requiring hundreds of portrait images and hours of training time,
to translate its generative space from the face domain to the artistic portrait domain.
It shows great superiority in quality, image resolution, data requirement, and efficiency compared to image style transfer and image-to-image translation models.

The strategy above, while effective, only learns an overall translation of the distribution, incapable of performing exemplar-based style transfer.
For a StyleGAN that has been transferred for generating a fixed caricature style, a laughing face will be largely mapped to its nearest one in the caricature domain, \ie, a portrait with an exaggerated mouth.
Users have no means of shrinking the face to pastiche their preferred artworks like in Fig.~\ref{fig:teaser}(c).
Although StyleGAN provides inherent exemplar-based single-domain style mixing by latent swapping~\cite{karras2019style,abdal2019image2stylegan},
such single-domain-oriented operation is counter-intuitive and incompetent for style transfer involving a source domain and a target domain.
This is because misalignment between these two domains may lead to unwanted artifacts during style mixing, especially for domain-specific structures.
However, importantly, a professional pastiche, should imitate how an artist handles face structures, \eg, abstraction in cartoons and deformation in caricatures.

To tackle these challenges, we propose a novel \textbf{DualStyleGAN} to realize effective modelling and control of dual styles for exemplar-based portrait style transfer.
DualStyleGAN retains an \textbf{intrinsic style path} of StyleGAN to control the style of the original domain,
while adding an \textbf{extrinsic style path} to model and control the style of the target extended domain, which naturally correspond to the content path and style path in the standard style transfer paradigm.
Moreover, the extrinsic style path inherits the hierarchical architecture from StyleGAN to modulate structural styles in coarse-resolution layers and color styles in fine-resolution layers for flexible multi-level style manipulations.

Adding an extrinsic style path to the original StyleGAN architecture is non-trivial for our task as it risks altering the generative space and behavior of the pre-trained StyleGAN.
To overcome this challenge, we present effective ways and insights to design the extrinsic style path and train DualStyleGAN.
1) \textit{Model design}: based on the analysis on the \ys{fine-tuning} behavior of StyleGAN, we propose to introduce the extrinsic style in a residual manner to the convolution layers,
which can well approximate how fine-tuning affects the convolution layers of StyleGAN.
We show that such design enables DualStyleGAN to effectively modulate the key structural styles.
2) \textit{Model training}: we introduce a novel progressive \ys{fine-tuning} methodology, where the extrinsic style path is first elaborately initialized so that DualStyleGAN retains the generative space of StyleGAN for seamless transfer learning. Then, we start out training DualStyleGAN with an easy style transfer task and then gradually increases the task difficulty, to progressively translate its generative space to the target domain.
In addition, we present a facial destylization method to provide face-portrait pairs, serving as supervision to promote the model to learn diverse styles and avoid mode collapse.

With the novel formulation above, the proposed DualStyleGAN offers high-quality and high-resolution pastiches and provides flexible and diverse control over both color styles and complicated structural styles, as shown in Fig.~\ref{fig:teaser}.
In summary, our contributions are threefold:
\begin{itemize}[itemsep=1.5pt,topsep=1pt,parsep=0pt]
  \item We propose a novel DualStyleGAN to characterize and control the intrinsic and extrinsic styles for exemplar-based high-resolution portrait style transfer, requiring only a few hundred style examples, which achieves superior performance over state-of-the-art methods in high-quality and diverse artistic portrait generation.
  \item We design a principled extrinsic style path to introduce style features from external domains via \ys{fine-tuning} and to provide hierarchical style manipulation in terms of both color and structure.
  \item We propose a novel progressive \ys{fine-tuning} scheme for robust transfer learning over networks with architecture modifications.
\end{itemize}

\section{Related Work}\vspace{-1mm}
\label{sec:related_work}

\noindent
\textbf{Artistic portrait generation with StyleGAN.}
StyleGAN~\cite{karras2019style,karras2020analyzing} synthesizes high-resolution face images with hierarchical style control.
Pinkney and Adler~\cite{pinkney2020resolution} fine-tuned StyleGAN on limited cartoon data, and found it promising in generating plausible cartoon faces.
The original model and fine-tuned model exhibit a reasonable degree of semantic alignment~\cite{wu2021stylealign}, allowing one to toonify a real face by applying its embedded latent code in the original model to the fine-tuned model to obtain the corresponding stylized face.
This framework is efficient and data-friendly, attracting further in-depth research, such as embedding acceleration~\cite{richardson2020encoding}, better choice of the latent code~\cite{song2021agilegan}, training on extremely limited data~\cite{jiang2021deceive,ojha2021few}.
In contrast to our work, these methods only learn an overall distribution translation without exemplar-based style control.
Kwong~\etal~\cite{kwong2021unsupervised} achieves style transfer by swapping fine-resolution-layer features from the exemplar style image with those from the content image under the assumption of model alignment.
However, the alignment gets weakened along with unconditional fine-tuning without valid supervision, eventually leading to a failure in layer swapping.
Thus, the method is mainly suitable for color transfer and not effective in controlling the vital structural styles.
By comparison, our model has an explicit extrinsic style path that can be conditionally trained to characterize the structural styles.
Moreover, supervision for learning diverse styles is provided via facial destylization.

\noindent
\textbf{Image-to-image translation.}
Portrait style transfer can be realized by an image-to-image translation framework~\cite{zhao2020unpaired,nizan2020breaking,xie2021unaligned,shao2021spatchgan}.
The main idea is to learn a bi-directional mapping~\cite{Zhu2017Unpaired} between the face and artistic portrait domains.
To find a correspondence between domains with a large appearance discrepancy,
U-GAT-IT~\cite{kim2019u} uses
attention modules to focus on key regions shared between domains.
AniGAN~\cite{li2021anigan} uses shared layers in discriminators to extract common features of two domains.
GNR~\cite{chong2021gans} learns effective content features and style features as those unchanged or altered during data augmentation, respectively.
For the style of caricatures, explicit image warping is applied to imitate the distinct facial deformation~\cite{cao2018carigans,shi2019warpgan}.
These strategies allow the image-to-image translation framework to stylize human faces involving drastic transformation.
However, learning from scratch on complex bi-directional translations makes this framework limited to low-resolution images and requires long training time.
Our method follows the \ys{fine-tuning} framework of StyleGAN, which is efficient in creating high-resolution portraits and provides flexible hierarchical style control beyond the capability of above methods.

\begin{figure}[t]
\centering
\includegraphics[width=0.97\linewidth]{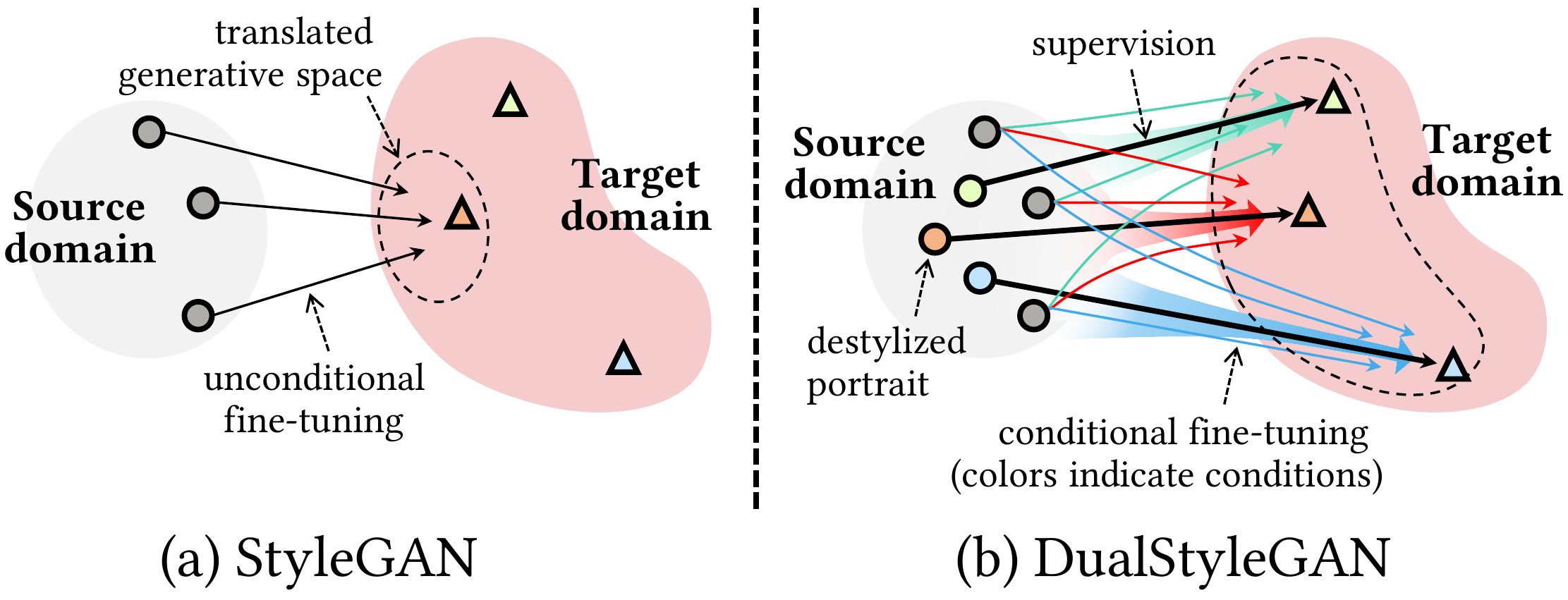}\vspace{-2mm}
\caption{Compare the unconditional \ys{fine-tuning} over StyleGAN and the conditional \ys{fine-tuning} over DualStyleGAN.}
\label{fig:distribution}\vspace{-1.5em}
\end{figure}

\section{Portrait Style Transfer via DualStyleGAN}

Our goal is to build DualStyleGAN based on a pre-trained StyleGAN, which can be transferred to a new domain and characterize the styles of both the original and the extended domains. Unconditional \ys{fine-tuning} translates the StyleGAN generative space as a whole, leading to the loss of diversity of the captured styles, as illustrated in Fig.~\ref{fig:distribution}.
Our key idea is to seek valid supervision to learn diverse styles (Sec.~\ref{sec:destylization}),
and to explicitly model the two kinds of styles with two individual style paths (Sec.~\ref{sec:network}).
We train DualStyleGAN with a principled progressive strategy for robust conditional \ys{fine-tuning} (Sec.~\ref{sec:progressive}).

\begin{figure}[t]
\centering
\includegraphics[width=0.95\linewidth]{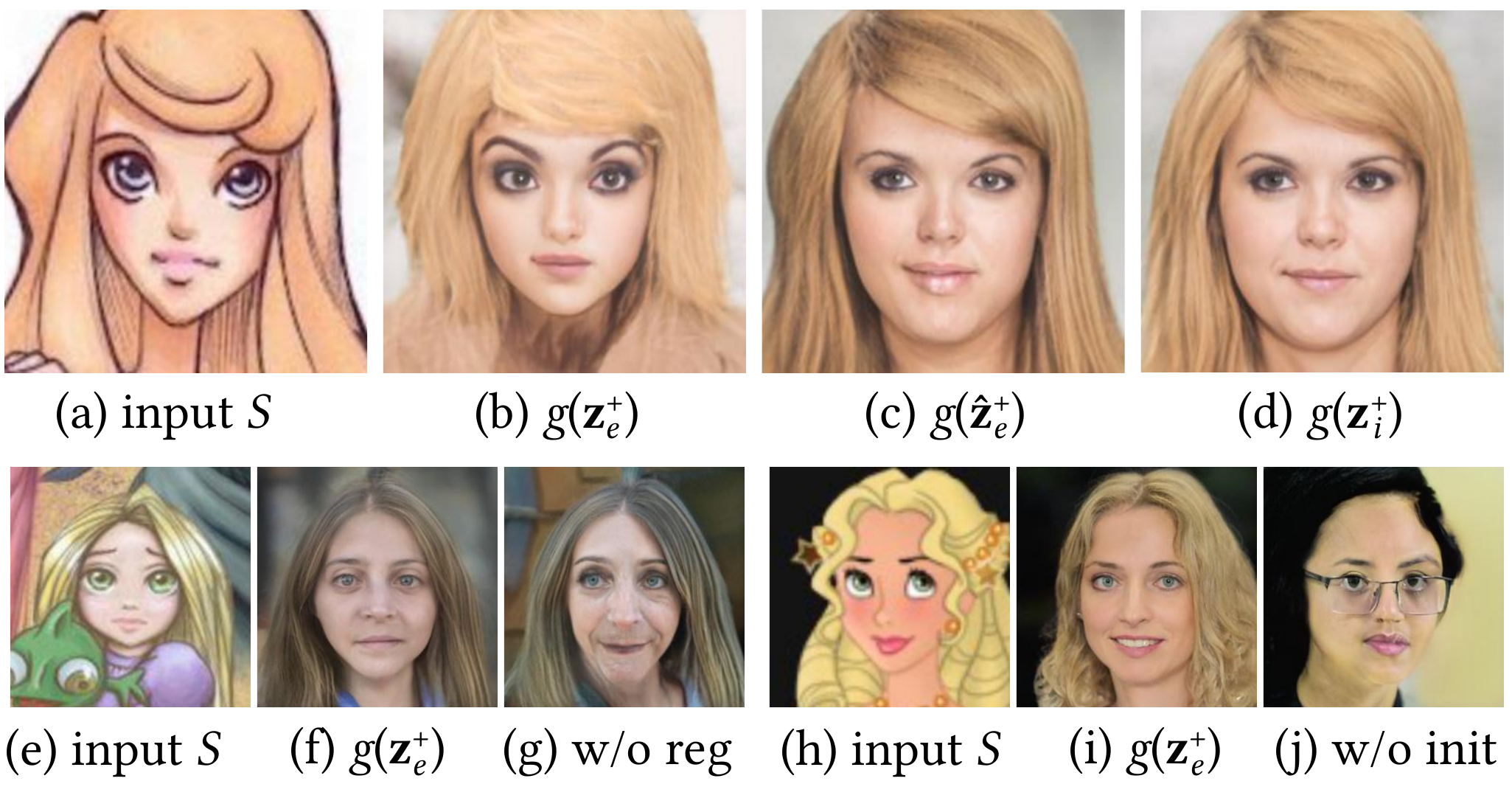}\vspace{-2.5mm}
\caption{Illustration of facial destylization. The destylized results of (a) in each stage are sequentially shown in (b)-(d) with the exaggerated eyes gradually turning realistic.
(e)-(g): Regularization prevents overfitting to the face-irrelevant green toy. (h)-(j): $\mathbf{z}^+_e$ serves as a good initial value to fit the complex cartoon faces.}
\label{fig:cartoontophoto}\vspace{-1.5em}
\end{figure}

\subsection{Facial Destylization}
\label{sec:destylization}

Facial destylization aims to recover realistic faces from artistic portraits to form anchored face-portrait pairs as supervision.
Given artistic portraits of the target domain, we would like to find their reasonable counterparts in the face domain.
Since the two domains might have a large appearance discrepancy,
it poses us a non-trivial challenge to balance between face realism and fidelity to the portraits.
To solve this problem, we propose a multi-stage destylization method to gradually enhance the realism of a portrait.

\noindent
\textbf{Stage I: Latent initialization}. The artistic portrait $S$ is first embeded into the StyleGAN latent space by an encoder $E$.
Here, we use a pSp encoder~\cite{richardson2020encoding} and modify it to embed FFHQ faces~\cite{karras2019style} into $\mathcal{Z}+$ space,
which is more robust to face-irrelevant background details and distorted shapes than the original $\mathcal{W}+$ space, as suggested in~\cite{song2021agilegan}.
An example of the reconstructed face $g(\mathbf{z}^+_e)$ is shown in Fig.~\ref{fig:cartoontophoto}(b), with $g$ the StyleGAN pre-trained on FFHQ and $\mathbf{z}^+_e=E(S)\in\mathbb{R}^{18\times512}$ the latent code.
Though $E$ is trained on real faces, $E(S)$ well captures the color and the structure of portrait $S$.

\noindent
\textbf{Stage II: Latent optimization}. In~\cite{pinkney2020resolution}, a face image is stylized by optimizing a latent code of $g$ to reconstruct this image~\cite{abdal2019image2stylegan} and applying this code to a fine-tuned model $g'$.
We take a reverse step to optimize the latent $\mathbf{z}^+$ of $g'$ to reconstruct $S$ with a novel regularization term, and apply the resulting $\hat{\mathbf{z}}^+_e$ to $g$ to obtain its destylized version,\vspace{-1mm}
\begin{equation}
\begin{aligned}
  \hat{\mathbf{z}}^+_e=&\arg\min_{\mathbf{z}^+} \mathcal{L}_{\text{perc}}(g'(\mathbf{z}^+), S)\\
  &+\lambda_{\text{ID}}\mathcal{L}_{\text{ID}}(g'(\mathbf{z}^+), S)+\|\sigma(\mathbf{z}^+)\|_1,\vspace{-1mm}
\end{aligned}
\end{equation}
where $\mathcal{L}_{\text{perc}}$ is the perceptual loss~\cite{Johnson2016Perceptual}, $\mathcal{L}_{\text{ID}}$ is the identity loss~\cite{deng2019arcface} to preserve the identity of the face and $\sigma(\mathbf{z}^+)$ is the standard error of 18 different 512-dimension vectors in $\mathbf{z}^+$. $\lambda_{\text{ID}}=0.1$.
Different from~\cite{abdal2019image2stylegan}, we design the regularization term to pull $\hat{\mathbf{z}}^+_e$ to the well-defined $\mathcal{Z}$ space to avoid overfitting as in Fig.~\ref{fig:cartoontophoto}(f)(g),
and use $\mathbf{z}^+_e$ rather than the mean latent code to initialize $\mathbf{z}^+$ before optimization, which helps accurately fit the face structures as in Fig.~\ref{fig:cartoontophoto}(i)(j).

\noindent
\textbf{Stage III: Image embedding}. Finally, we embed $g(\hat{\mathbf{z}}^+_e)$ as $\mathbf{z}^+_i=E(g(\hat{\mathbf{z}}^+_e))$, which further eliminates unreal facial details. The resulting $g(\mathbf{z}^+_i)$ has reasonable facial structures, providing valid supervision on how to deform and abstract the facial structures to imitate $S$.

\begin{figure}[t]
\centering
\includegraphics[width=0.97\linewidth]{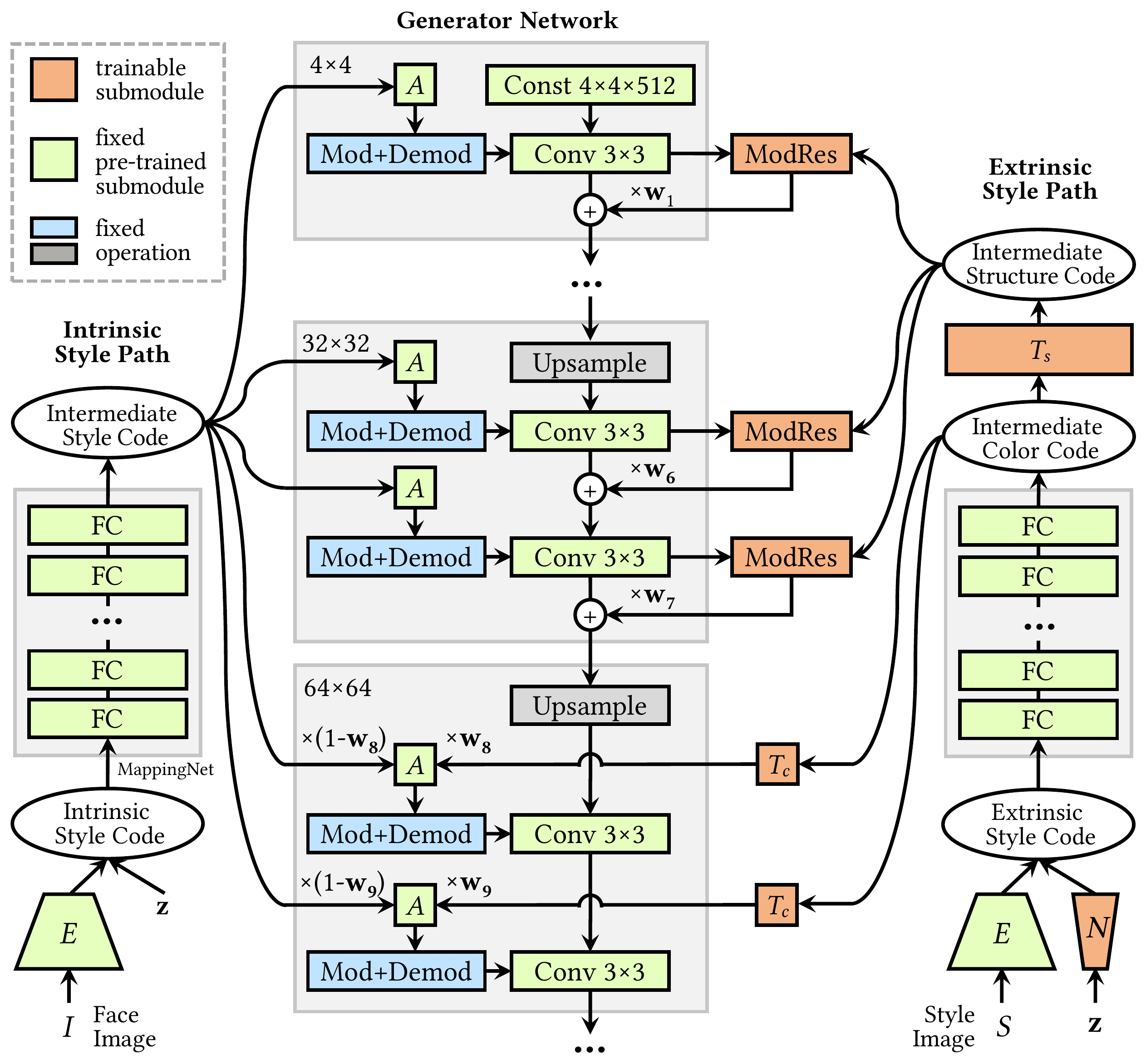}\vspace{-1mm}
\caption{Network details of DualStyleGAN. For simplicity, the learned weights, biases and noises of StyleGAN are omitted.}\vspace{-1.6em}
\label{fig:network}
\end{figure}

\subsection{DualStyleGAN}
\label{sec:network}

Figure~\ref{fig:network} shows the network details of DualStyleGAN $G$.
The intrinsic style path and generator network form a standard StyleGAN and are kept fixed during \ys{fine-tuning}.
The intrinsic style path accepts intrinsic style code of unit Gaussian noise $\mathbf{z}\in\mathbb{R}^{1\times512}$, $\mathbf{z}^+_i$ of artistic portraits or $\mathbf{z}^+$ of real faces embedded by $E$.
The extrinsic style path simply uses $\mathbf{z}^+_e$ of artistic portraits as the extrinsic style code, which
captures meaningful semantic cues like hair colors and facial shapes (Fig.~\ref{fig:cartoontophoto}(b)).
Extrinsic style codes can also be sampled via a sampling network $N$ by mapping unit Gaussian noises to the extrinsic style distribution.
Formally, given a face image $I$ and an artistic portrait image $S$, exemplar-based style transfer is achieved by $G(E(I),E(S),\mathbf{w})$, where $\mathbf{w}\in\mathbb{R}^{18}$ is the weight vector for flexible style combination of two paths, and is set to $\mathbf{1}$ by default.
Artistic portrait generation is realized by $G(\mathbf{z}_1,N(\mathbf{z}_2),\mathbf{w})$.
When $\mathbf{w}=\mathbf{0}$, $G$ degrades into a standard $g$ for face generation, \ie, $G(\mathbf{z},\cdot,\mathbf{0})\sim g(\mathbf{z})$.  %

StyleGAN provides a hierarchical style control, where fine-resolution and coarse-resolution layers model the low-level color style and high-level shape style, respectively, which inspires our design of the extrinsic style path.

\noindent
\textbf{Color control}. In fine-resolution layers (8$\sim$18), the extrinsic style path takes the same strategy as StyleGAN.
Specifically, $\mathbf{z}^+_e$ goes through a mapping network $f$, color transform blocks $T_c$ and affine transform blocks $A$.
The resulting style bias is fused with the style bias from the intrinsic style path with weight $\mathbf{w}$ for the final AdaIN~\cite{huang2017adain}.
Different from $g$, trainable $T_c$ composed of fully connected layers is added to characterize domain-specific colors.

\noindent
\textbf{Structure control}. In coarse-resolution layers (1$\sim$7), we propose modulative residual blocks (ModRes) to adjust structural styles and add a structure transform block $T_s$ to characterize domain-specific structural styles.
ModRes contains a ResBlock~\cite{he2016deep} to simulate the changes of convolution layers during fine-tuning and an AdaIN block for style condition.
To understand the motivation of the proposed ModRes, below we provide some experimental analysis of the \ys{fine-tuning} behavior over StyleGAN.

\begin{figure}[t]
\centering
\includegraphics[width=0.98\linewidth]{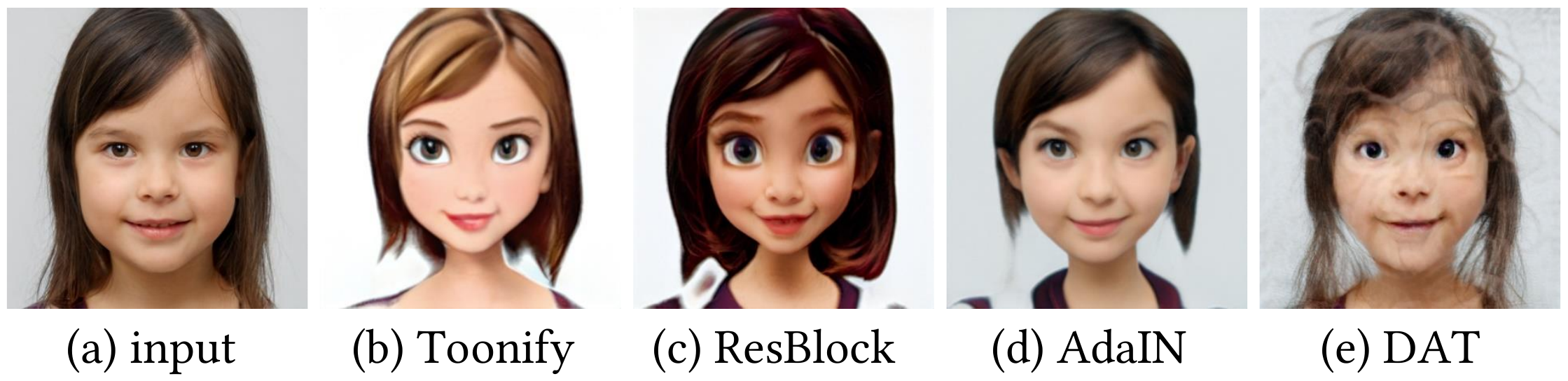}\vspace{-1.5mm}
\caption{ResBlocks best simulate Toonify~\cite{pinkney2020resolution}.}
\label{fig:path}\vspace{-1.7em}
\end{figure}

\noindent
\textbf{Simulating \ys{fine-tuning} behavior}.
The success of toonification~\cite{pinkney2020resolution} relies on the semantic alignment of the models before and after fine-tuning, namely, two models have shared latent spaces~\cite{kwong2021unsupervised} and closely-related convolution features. It also implies that the difference of these features is also closely-related to the original features. Moreover, among all submodules of StyleGAN, the convolution layers change the most during fine-tuning~\cite{wu2021stylealign}. Therefore, it is possible to keep all other submodules fixed but only learn changes over the convolution features to simulate the changes of the convolution weight matrices in \ys{fine-tuning}.
In StyleGAN, common adjustments over deep features involve channel-wise, spatial-wise and element-wise modulations, corresponding to AdaIN~\cite{huang2017adain}, Diagonal Attention (DAT)~\cite{kwon2021diagonal} and ResBlock, respectively.
We conduct a toy experiment and find that modulations in channel (Fig.~\ref{fig:path}(d)) or spatial (Fig.~\ref{fig:path}(e)) dimension alone are not enough to approximate the \ys{fine-tuning} behavior. ResBlocks achieve the most similar results (Fig.~\ref{fig:path}(c)) to those by fine-tuning the whole StyleGAN (Fig.~\ref{fig:path}(b)).
Therefore, we choose residual blocks and apply AdaIN to the convolution layers in the residual path to provide extrinsic style conditions.

\noindent
\textbf{Summary}.~Our DualStyleGAN is very simple yet effective. \textbf{1) Hierarchical modelling of complex styles}: It provides hierarchical modelling on both color and complex structural styles.
\textbf{2) Flexible style manipulation}: It supports flexible style mixing between two domains with weight $\mathbf{w}$.
\textbf{3) Alleviate mode collapse}: \ys{Fine-tuning} trains only the extrinsic style path while keeping the pre-trained StyleGAN intact, which preserves the original diverse facial features to avoid mode collapse.
\textbf{4) Structure preservation}: The additive property of our modulative residual block leads to a robust content loss,
as we will detail in Sec.~\ref{sec:progressive}.

\begin{figure}[t]
\centering
\includegraphics[width=0.97\linewidth]{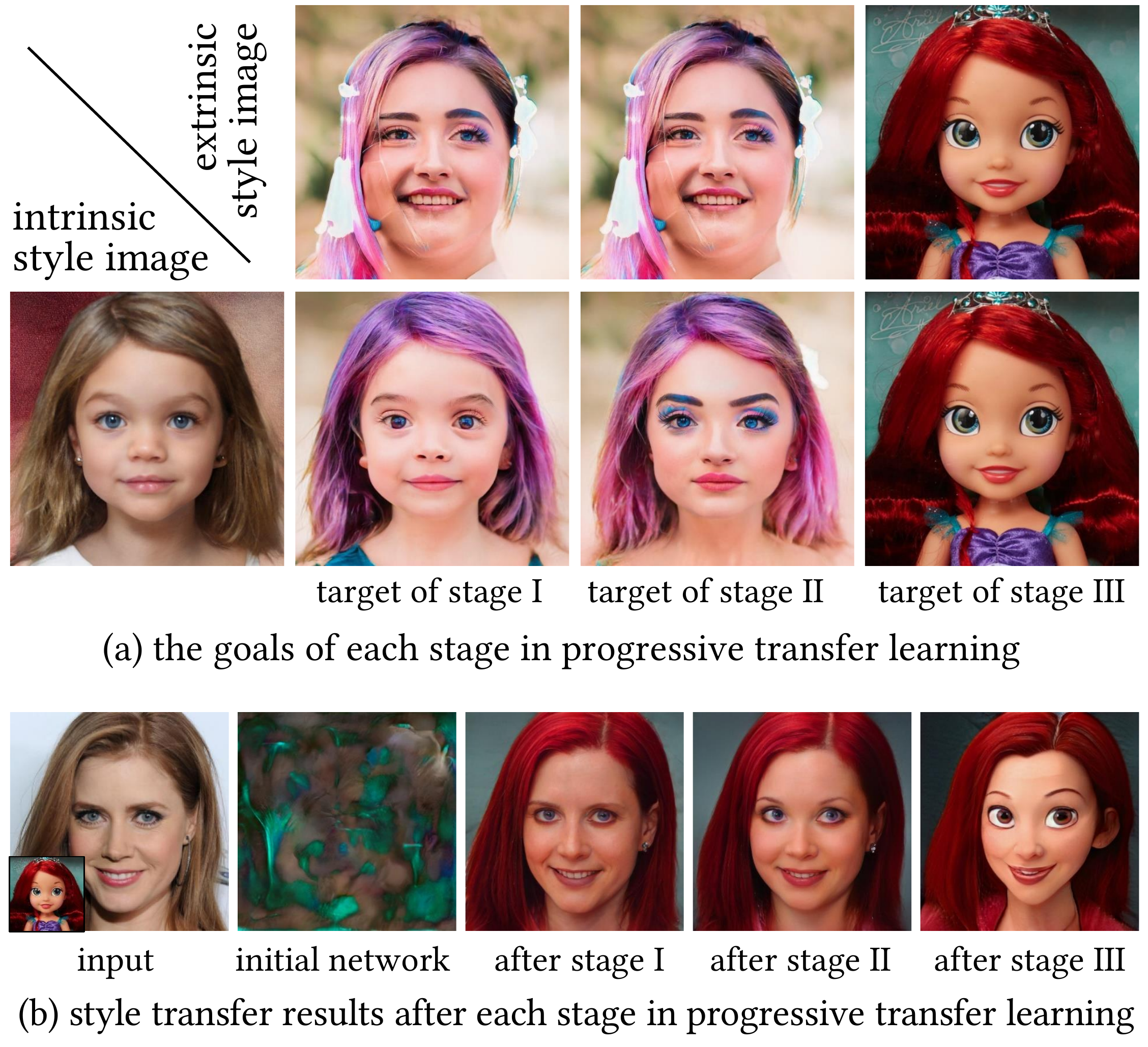}\vspace{-2.5mm}
\caption{Illustration of progressive \ys{fine-tuning}. (a) DualStyleGAN is tasked with style transfers with growing difficulties. (b) The performance of DualStyleGAN after each stage.}\vspace{-0.5em}
\label{fig:pretrain}
\end{figure}

\begin{figure}[t]
\centering
\includegraphics[width=0.97\linewidth]{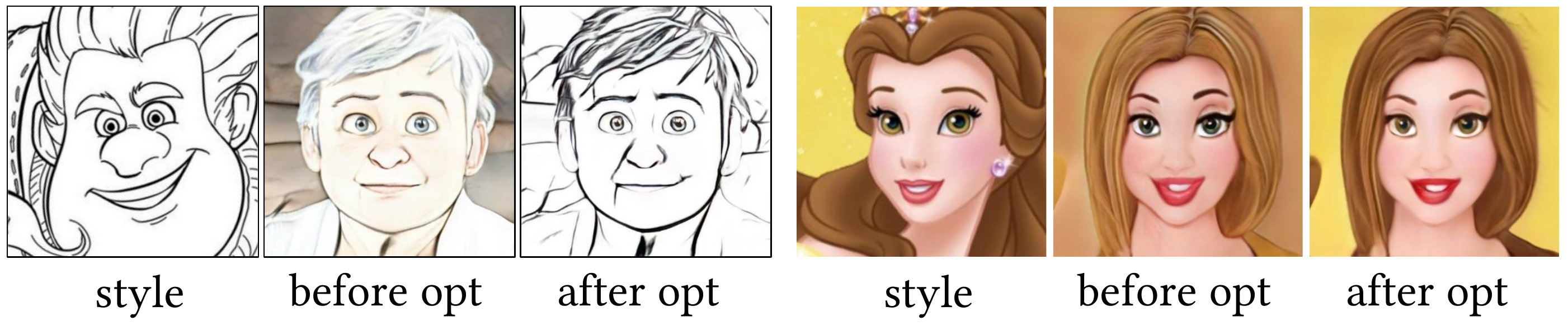}\vspace{-2mm}
\caption{Optimize the extrinsic style code to refine color.}\vspace{-1.7em}
\label{fig:latentoptimization}
\end{figure}

\subsection{Progressive \ys{Fine-Tuning}}
\label{sec:progressive}

We propose a progressive \ys{fine-tuning} scheme to smoothly transform the generative space of DualStyleGAN towards the target domain.
The scheme borrows the idea of curriculum learning~\cite{bengio2009curriculum} to gradually increase the task difficulty in three stages as illustrated in Fig.~\ref{fig:pretrain}(a).

\noindent
\textbf{Stage I: Color transfer on source domain}. DualStyleGAN is tasked with color transfer within the source domain in this stage. Thanks to the design of our extrinsic style path, it can be achieved purely by a specific model initialization. Specifically, the convolution filters in the modulative residual blocks are set to values close to 0 in order to produce negligible residual features and the fully connected layers in color transform blocks are initialized with identity matrices, meaning no changes to the input latent code.
To this end, DualStyleGAN runs the standard style mixing operation of StyleGAN,
where fine-resolution and coarse-resolution layers use the latent codes from the intrinsic and extrinsic style paths, respectively. As shown in Fig.~\ref{fig:pretrain}(b), the initialized DualStyleGAN generates plausible human faces that still lie in the generative space of the pre-trained StyleGAN, allowing smooth fine-tuning in the next stage.

\noindent
\textbf{Stage II: Structure transfer on source domain}. This stage aims to fine-tune DualStyleGAN on the source domain to fully train its extrinsic style path to capture and transfer mid-level styles. StyleGAN's style mixing in middle layers involves small-scale style transfer like makeups, which provides DualStyleGAN with effective supervision. In stage II, we draw random latent code $\mathbf{z}_1$ and $\mathbf{z}_2$, and would like $G(\mathbf{z}_1,\mathbf{\tilde{z}}_2,\mathbf{1})$ to approximate the style mixing target  $g(\mathbf{z}^+_l)$ with perceptual loss, where $\mathbf{\tilde{z}}_2$ is sampled from $\{\mathbf{z}_2, E(g(\mathbf{z}_2))\}$, $l$ is the layer where style mixing occurs and $\mathbf{z}^+_l\in \mathcal{Z}+$ is a concatenation of $l$ vector $\mathbf{z}_1$ and (18$-l$) vector $\mathbf{z}_2$. We gradually decrease $l$ from $7$ to $5$ during fine-tuning with the following objective:\vspace{-1mm}
\begin{equation}
  \min_{G}\max_{D}\lambda_{\text{adv}}\mathcal{L}_{\text{adv}}+\lambda_{\text{perc}}\mathcal{L}_{\text{perc}}(G(\mathbf{z}_1,\mathbf{\tilde{z}}_2,\mathbf{1}),g(\mathbf{z}^+_l)),\vspace{-1mm}
\end{equation}
where $\mathcal{L}_{\text{adv}}$ is the StyleGAN adversarial loss.
\ys{By decreasing $l$, $g(\mathbf{z}^+_l)$ will have more structure styles from $\mathbf{\tilde{z}}_2$. Thus, the extrinsic style path will learn to capture and transfer more structure styles besides colors.}

\noindent
\textbf{Stage III: Style transfer on target domain}.
Finally, we fine-tune DualStyleGAN on the target domain. We would like the style codes $\mathbf{z}^+_i$ and $\mathbf{z}^+_e$ of an exemplar artistic portrait $S$ to reconstruct $S$ with $\mathcal{L}_{\text{perc}}(G(\mathbf{z}^+_i,\mathbf{z}^+_e,\mathbf{1}), S)$.
As in the standard exemplar-based style transfer paradigm, for a random intrinsic style code $\mathbf{z}$, we apply style loss\vspace{-1mm}
\begin{equation}
  \mathcal{L}_{\text{sty}}=\lambda_{\text{CX}}\mathcal{L}_{\text{CX}}(G(\mathbf{z},\mathbf{z}^+_e,\mathbf{1}),S)
  +\lambda_{\text{FM}}\mathcal{L}_{\text{FM}}(G(\mathbf{z},\mathbf{z}^+_e,\mathbf{1}),S),\nonumber\vspace{-1mm}
\end{equation}
where $\mathcal{L}_{\text{CX}}$ is the contextual loss~\cite{mechrez2018contextual} and $\mathcal{L}_{\text{FM}}$ is the feature matching loss~\cite{huang2017adain}, to match the style of $G(\mathbf{z},\mathbf{z}^+_e,\mathbf{1})$ to $S$.
For content loss, we use the identity loss~\cite{deng2019arcface} and $L_2$ regularization of the weight matrices of ModRes,\vspace{-1mm}
\begin{equation}
\label{eq:content_loss}
  \mathcal{L}_{\text{con}}=\lambda_{\text{ID}}\mathcal{L}_{\text{ID}}(G(\mathbf{z},\mathbf{z}^+_e,\mathbf{1}),g(\mathbf{z}))+\lambda_{\text{reg}}\|W\|_2.\vspace{-1mm}
\end{equation}
Similar to the initialization in Stage I, regularization over weight matrices makes the residual features close to zeros, which preserves the original intrinsic facial structures and prevents overfitting. Our full objectives take the form of\vspace{-1mm}
\begin{equation}\label{eq:total_loss}
  \min_{G}\max_{D}\lambda_{\text{adv}}\mathcal{L}_{\text{adv}}+\lambda_{\text{perc}}\mathcal{L}_{\text{perc}}+\mathcal{L}_{\text{sty}}+\mathcal{L}_{\text{con}}.\vspace{-1mm}
\end{equation}

\subsection{Latent Optimization and Sampling}\vspace{-1mm}
\label{sec:post}

\noindent
\textbf{Latent optimization.}
It is hard to fully capture the extremely diverse styles.
To solve this problem, we fix DualStyleGAN and optimize each extrinsic style code to fit its ground truth $S$.
The optimization follows the process of embedding an image into the latent space~\cite{abdal2019image2stylegan} and minimizes a perceptual loss and a contextual loss in Eq.~(\ref{eq:total_loss}). As shown in Fig.~\ref{fig:latentoptimization}, the colors are well refined by latent optimization.

\begin{figure*}[t]
\centering
\includegraphics[width=0.97\linewidth]{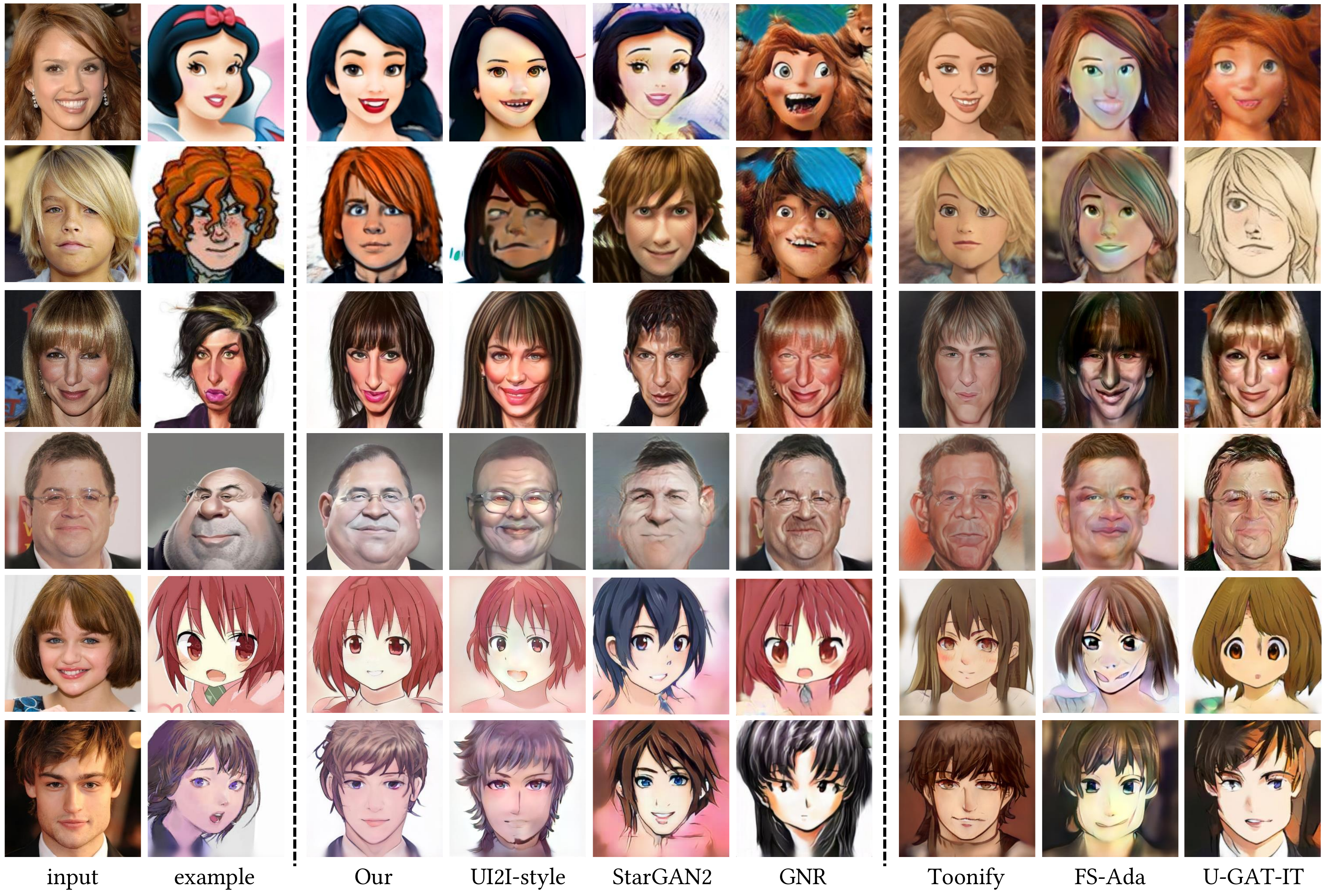}\vspace{-2mm}
\caption{Visual comparison on exemplar-based portrait style transfer.}\vspace{-1em}
\label{fig:comparison}
\end{figure*}

\noindent
\textbf{Latent sampling.} To sample random extrinsic styles, we train a sampling network $N$ to map unit Gaussian noises to the distribution of optimized extrinsic style codes using a maximum likelihood criterion~\cite{hoshen2019non}. Please refer to~\cite{hoshen2019non} for the details. Since structures (first 7 rows of $\mathbf{z}^+_e$) and colors (last 11 rows of $\mathbf{z}^+_e$) are well disentangled in DualStyleGAN, we treat these two parts separately, \ie, structure code and color code are independently sampled from $N$ and concatenated to form the complete extrinsic style code.

\vspace{-1mm}
\section{Experiments}\vspace{-2mm}
\label{sec:experiment}

\noindent
\textbf{Datasets.}~Our goal is to allow users to collect portrait images of their favourite artworks for DualStyleGAN to pastiche.~We would like the dataset to be limited to a few hundred images for easy collection. Therefore, we choose three datasets in popular styles of cartoon, caricature, anime.~Cartoon dataset~\cite{pinkney2020resolution} has 317 images.
We use 199 images from WebCaricature~\cite{Huo2017Variation,Huo2018WebCaricature} and 174 images from Danbooru Portraits~\cite{danbooru2019Portraits} to build the Caricature and Anime datasets, respectively.
We test on the same datasets and CelebA-HQ~\cite{liu2015deep, karras2018progressive} for extrinsic and intrinsic styles, respectively.

\noindent
\textbf{Implementation details.} Our progressive \ys{fine-tuning} uses eight NVIDIA Tesla V100 GPUs and a batch size of $4$ per GPU. Stage II uses $\lambda_{\text{adv}}=0.1,\lambda_{\text{perc}}=0.5$, and trains on $l=7$, $6$, $5$ for 300, 300, 3000 iterations, respectively, taking about 0.5 hour.
Stage III sets $\lambda_{\text{adv}}=1,\lambda_{\text{perc}}=1,\lambda_{\text{CX}}=0.25,\lambda_{\text{FM}}=0.25$, sets $(\lambda_{\text{ID}}, \lambda_{\text{reg}})$ to $(1,0.015)$, $(4,0.005)$, $(1,0.02)$ and trains for 1400, 1000, 2100 iterations on cartoon, caricature and anime, respectively.
Training takes about 0.75 hour on average.
Destylization (Sec.~\ref{sec:destylization}), latent optimization and training sampling network (Sec.~\ref{sec:post}) use one GPU and take about 5, 1, 0.13 hours, respectively.
\ys{Testing takes about 0.13s per image.}
For simplicity, we use $[n_1*v_1, n_2*v_2, ...]$ to indicate the first $n_1$ weights in vector $\mathbf{w}$ are set to the value of $v_1$, the next $n_2$ weights are set to the value of $v_2$.  $\mathbf{w}_s$ and $\mathbf{w}_c$ denote the structure weight vector (the first $7$ weights of $\mathbf{w}$) and color weight vector (the last $11$ weights), respectively. By default, we set $\mathbf{w}$ to $\mathbf{1}$ for training and set $\mathbf{w}_c$ to $\mathbf{1}$, $\mathbf{w}_s$ to $\mathbf{0.75}$, $\mathbf{1}$ and $[4*0,3*0.75]$ for testing for cartoon, caricature, anime, respectively.

\subsection{Comparison with State-of-the-Art Methods}\vspace{-2mm}
\label{sec:comparison}

Figure~\ref{fig:comparison} presents the qualitative comparison with six state-of-the-art methods: image-to-image-translation-based StarGAN2~\cite{choi2020stargan}, GNR~\cite{chong2021gans}, U-GAT-IT~\cite{kim2019u}, and StyleGAN-based UI2I-style~\cite{kwong2021unsupervised}, Toonify~\cite{pinkney2020resolution}, Few-Shot Adaptation (FS-Ada)~\cite{ojha2021few}.
The image-to-image translation and FS-Ada use $256\times 256$ images.~Other methods support $1024\times 1024$.
Toonify, FS-Ada and U-GAT-IT learn domain-level rather than image-level styles. Thus their results are not consistent with the style examples.
The severe data imbalance problem makes it hard to train valid cycle translations. Thus, StarGAN2 and GNR overfit the style images and ignore the input faces on the anime style.
UI2I-style captures good color styles via layer swapping, but the model misalignment makes the structure features hard to blend, leading to failure structural style transfer, as also analyzed in Sec.~\ref{sec:related_work}.
By comparison, DualStyleGAN transfers the best style of the exemplar style in both colors and complex structures.

\begin{table} [t]
\caption{User preference scores. Best scores are marked in bold.}\vspace{-3mm}
\label{tb:user_study}
\centering
\footnotesize
\begin{tabular}{l|c|c|c|c}
\toprule
Method & Cartoon & Caricature & Anime & Average \\
\midrule
GNR~\cite{chong2021gans} & 0.01 & 0.06 & 0.04 & 0.04 \\
StarGANv2~\cite{choi2020stargan} & 0.01 & 0.00 & 0.04 & 0.02 \\
UI2I-style~\cite{kwong2021unsupervised} & 0.05 & 0.15 & 0.14 & 0.11 \\
Our & \textbf{0.93} & \textbf{0.79} & \textbf{0.78} & \textbf{0.83} \\
\bottomrule
\end{tabular}\vspace{-0.8em}
\end{table}

To quantitatively evaluate the performance, we conduct a user study, where 27 subjects are invited to select what they consider to be the best results from the four exemplar-based style transfer methods. Each style dataset uses 10 results for evaluation.
Table~\ref{tb:user_study}  summarizes the average preference scores, where DualStyleGAN receives the best score.

\begin{figure}[t]
\centering
\includegraphics[width=0.97\linewidth]{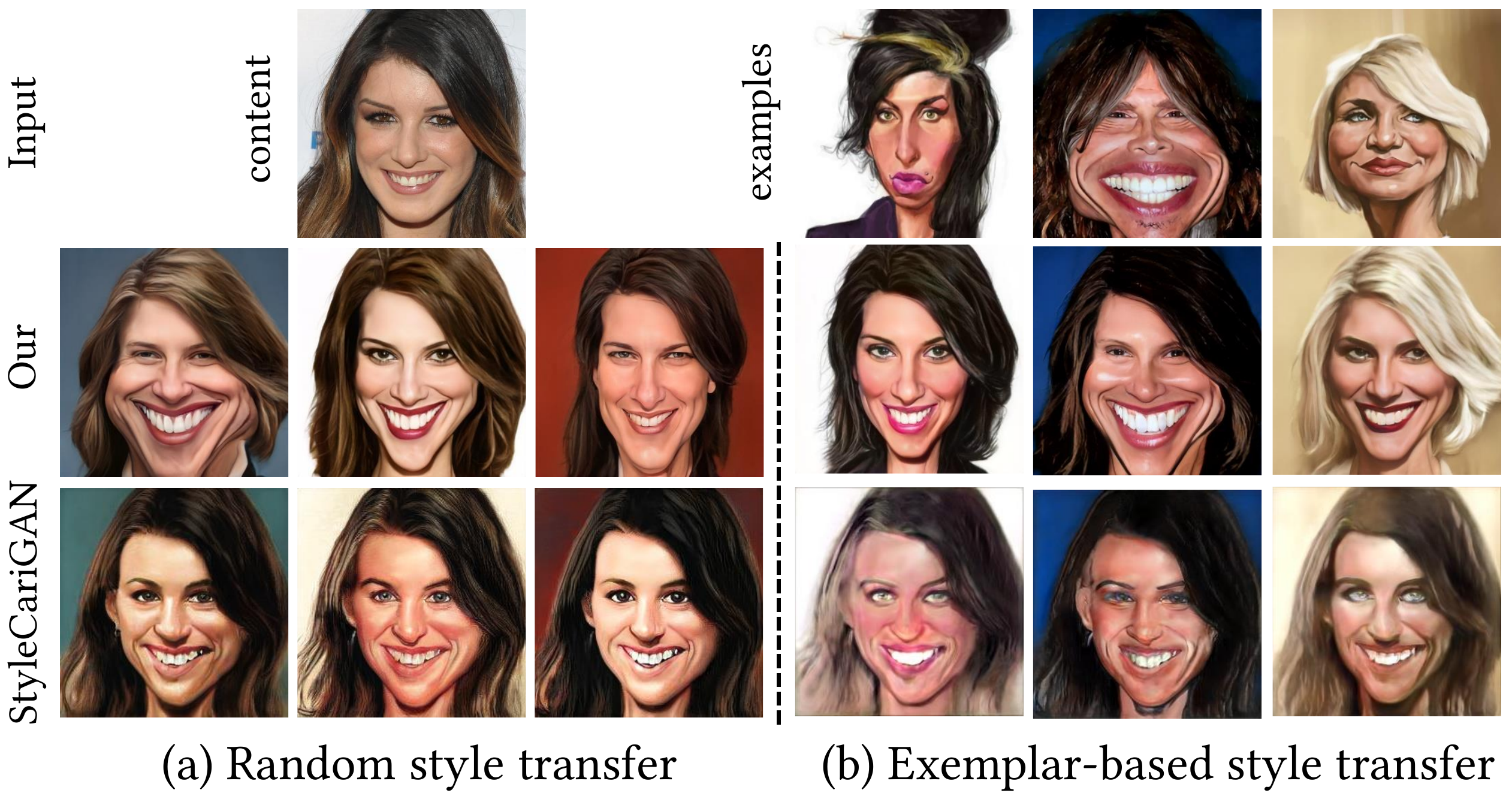}\vspace{-2.5mm}
\caption{Comparison with StyleCariGAN~\cite{jang2021stylecarigan}.}\vspace{-1.5em}
\label{fig:stylecarigan}
\end{figure}

\noindent
\ys{\textbf{Compare with StyleCariGAN.} We further compare with the advanced StyleCariGAN~\cite{jang2021stylecarigan} on caricatures.
StyleCariGAN combines StyleGAN and cycle translation to employ style mixing for color transfer and learn structure transfer with cycle translation.
We follow it to find latent codes of the content and example images via optimization~\cite{karras2020analyzing,tewari2020pie} for its inputs.
Depending on whether the latent codes are randomly sampled from the official style palette or from example images, StyleCariGAN can transfer random or exemplar-based styles on $256\times 256$ images.
As shown in Fig.~\ref{fig:stylecarigan}, StyleCariGAN generates the same facial structures since its cycle translation only learns an overall structure style.
By comparison, our method effectively adjusts the structure style based on the example.
Morevoer, our results are of higher resolution and visual quality, even if StyleCariGAN uses 6K training images, much more than ours.}

\subsection{Ablation Study}
\label{sec:ablation}

\begin{figure}[t]
\centering
\includegraphics[width=0.97\linewidth]{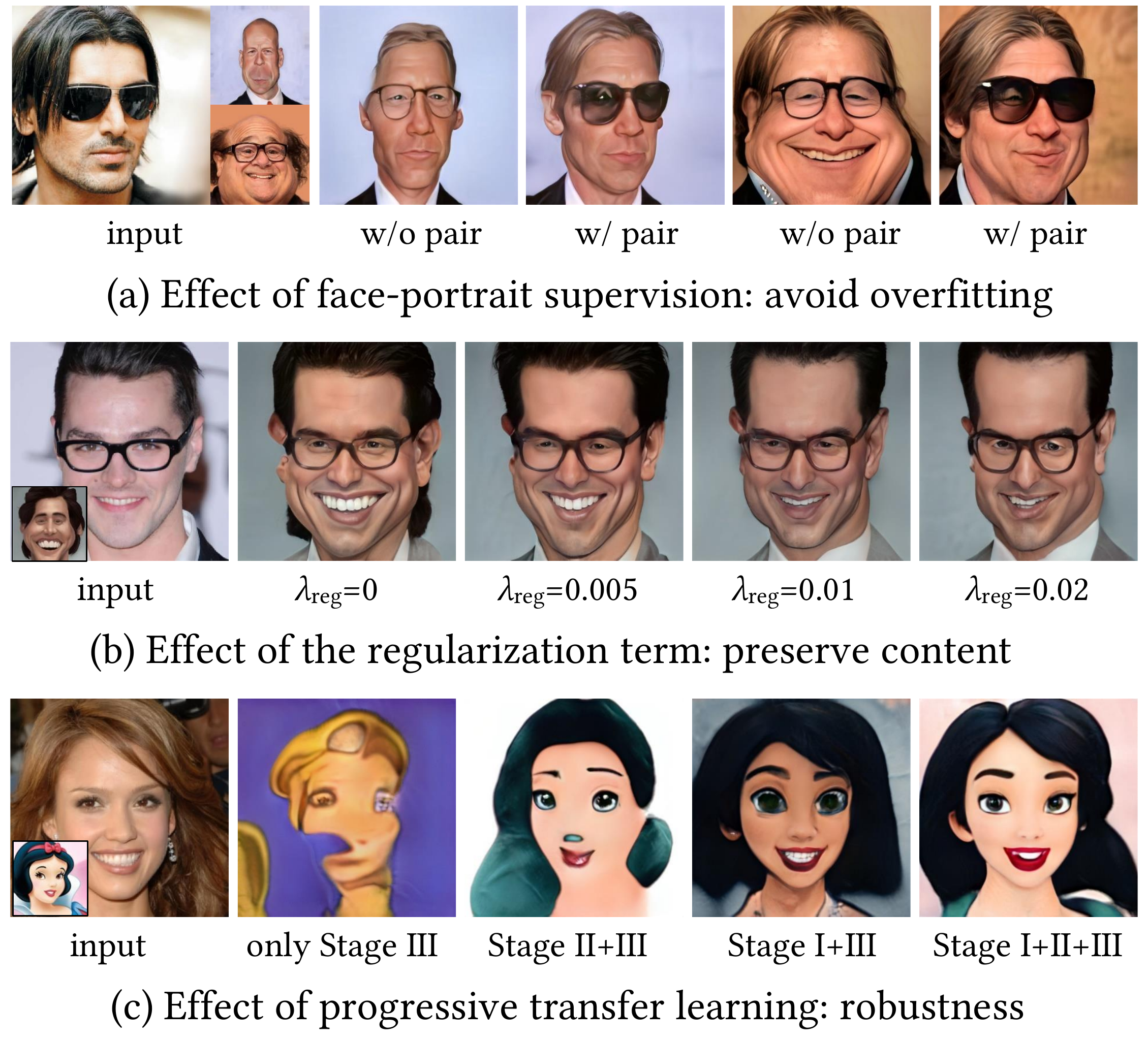}\vspace{-2.8mm}
\caption{Ablation study.}\vspace{-1em}
\label{fig:ablation}
\end{figure}

\noindent
\textbf{Paired data.} Figure~\ref{fig:ablation}(a) compares the results with and without face-portrait supervision in Sec.~\ref{sec:destylization}.
Without supervision, the model overfits the portraits without considering the input face structures.
The supervision effectively guides the model to find the structural relationship between the face and portrait, leading to more reasonable results.

\noindent
\textbf{Regularization.} The effect of the regularization term in the content loss (Eq.~(\ref{eq:content_loss})) is shown in Fig.~\ref{fig:ablation}(b).
Without the regularization term, the model overfits the hair styles of the example. Using the regularization term solves this issue. A large $\lambda_{\text{reg}}$ will over-preserve the input face shape like the mouth. Therefore, we use $\lambda_{\text{reg}}=0.005$ as a trade-off.

\noindent
\textbf{Progressive \ys{fine-tuning}.}~As shown in Fig.~\ref{fig:ablation}(c), without the initialization of Stage I, the generative space of the pre-trained StyleGAN is severely altered (Fig.~\ref{fig:pretrain}(b)), failing the transfer training completely.
Without pre-training on real faces to capture face semantic features, the extrinsic style path cannot fulfill the complex task in Stage III.
Only through the full progressive \ys{fine-tuning}, DualStyleGAN can accurately transfer the extrinsic styles.

\begin{figure}[t]
\centering
\includegraphics[width=0.97\linewidth]{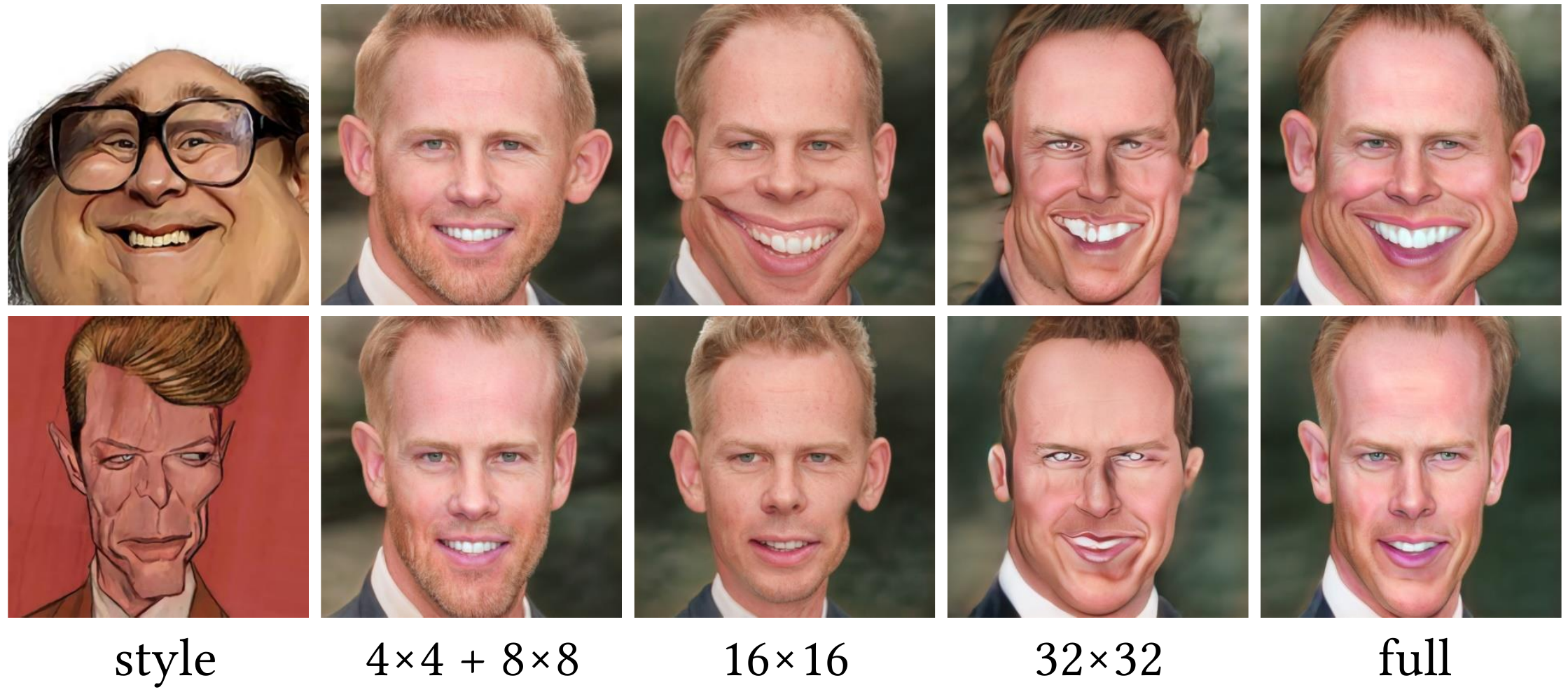}\vspace{-2mm}
\caption{The proposed extrinsic style path learns semantically hierarchical structure modulations.}\vspace{-1.5em}
\label{fig:deformation}
\end{figure}

\noindent
\textbf{Effect of different layers.} To study how each layer of the extrinsic style path affects the facial features, each time we activate a subset of layers (for example, $\mathbf{w}=[3*0,2*1,13*0]$ only activates two $16\times16$ layers) and compare their results in Fig.~\ref{fig:deformation}. Since the AdaIN-based color modulation has been well studies in StyleGAN, we only focus on the structure modulation in coarse-resolution layers.
It can be seen that the initial layers adjust the overall face shapes, $16\times16$ layers exaggerate facial components like mouths, and $32\times32$ layers focus on local shapes like wrinkles.

\subsection{Further Analysis}
\label{sec:more_result}

\begin{figure}[t]
\centering
\includegraphics[width=0.97\linewidth]{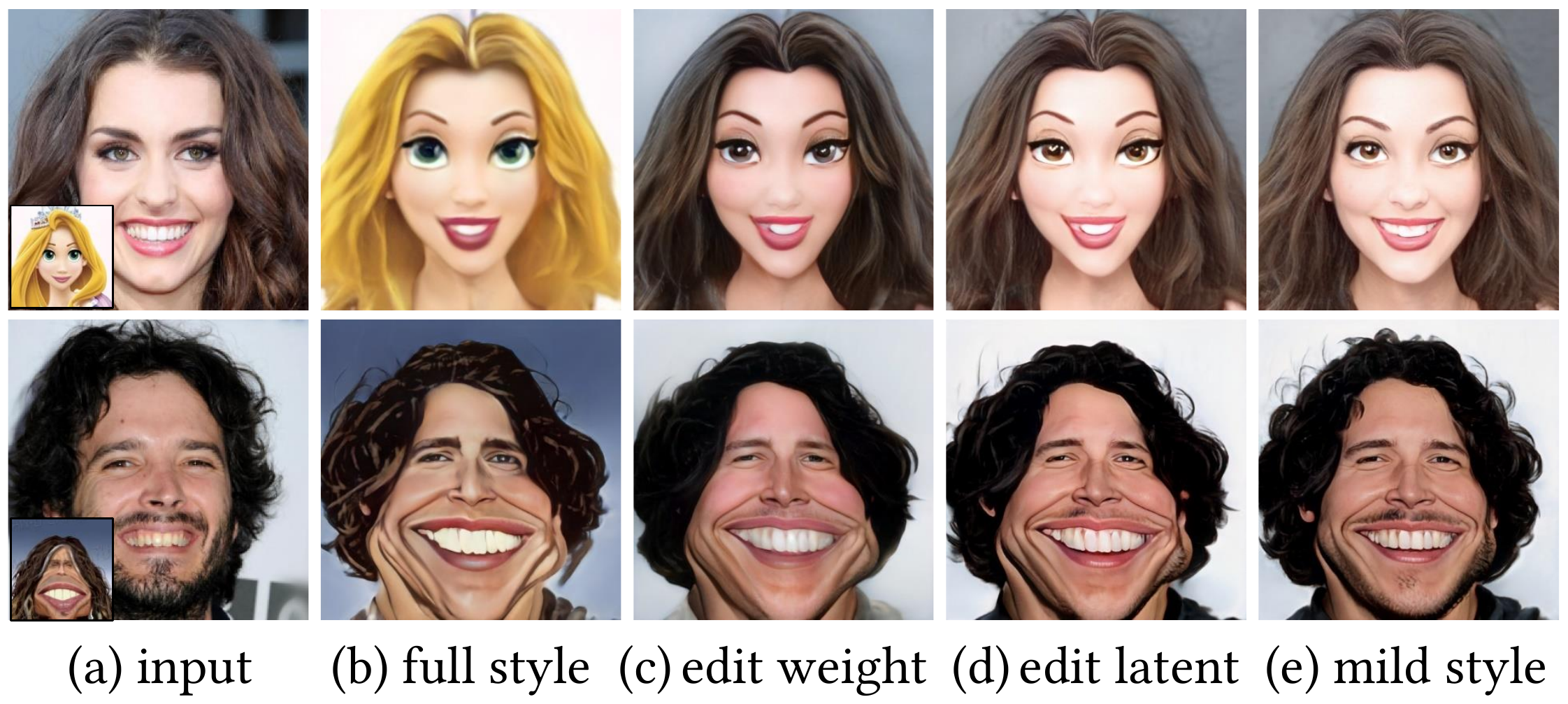}\vspace{-2mm}
\caption{Preservation of color and structure from the photo.}\vspace{-1em}
\label{fig:colorpreserve}
\end{figure}

\noindent
\textbf{Color and structure preservation.} Users may want to keep the color of the original photo as Toonify~\cite{pinkney2020resolution}.
We provide two ways of color preservation.
The first is just to deactivate color-related layers in the extrinsic style path by setting $\mathbf{w}_c=\mathbf{0}$ as in Fig.~\ref{fig:colorpreserve}(c).
Another way is to replace the extrinsic style codes with intrinsic style codes in the last 11 layers.
Compared to the first way, the intrinsic latent codes additionally go through color transform blocks, making the final color more aligned with the target domain as in Fig.~\ref{fig:colorpreserve}(d).
Finally, structure preservation can be easily achieved by setting $\mathbf{w}_s<\mathbf{1}$. Figure~\ref{fig:colorpreserve}(e) presents examples of mild style transfer with $\mathbf{w}_s=\mathbf{0.5}$.

\begin{figure}[t]
\centering
\includegraphics[width=0.97\linewidth]{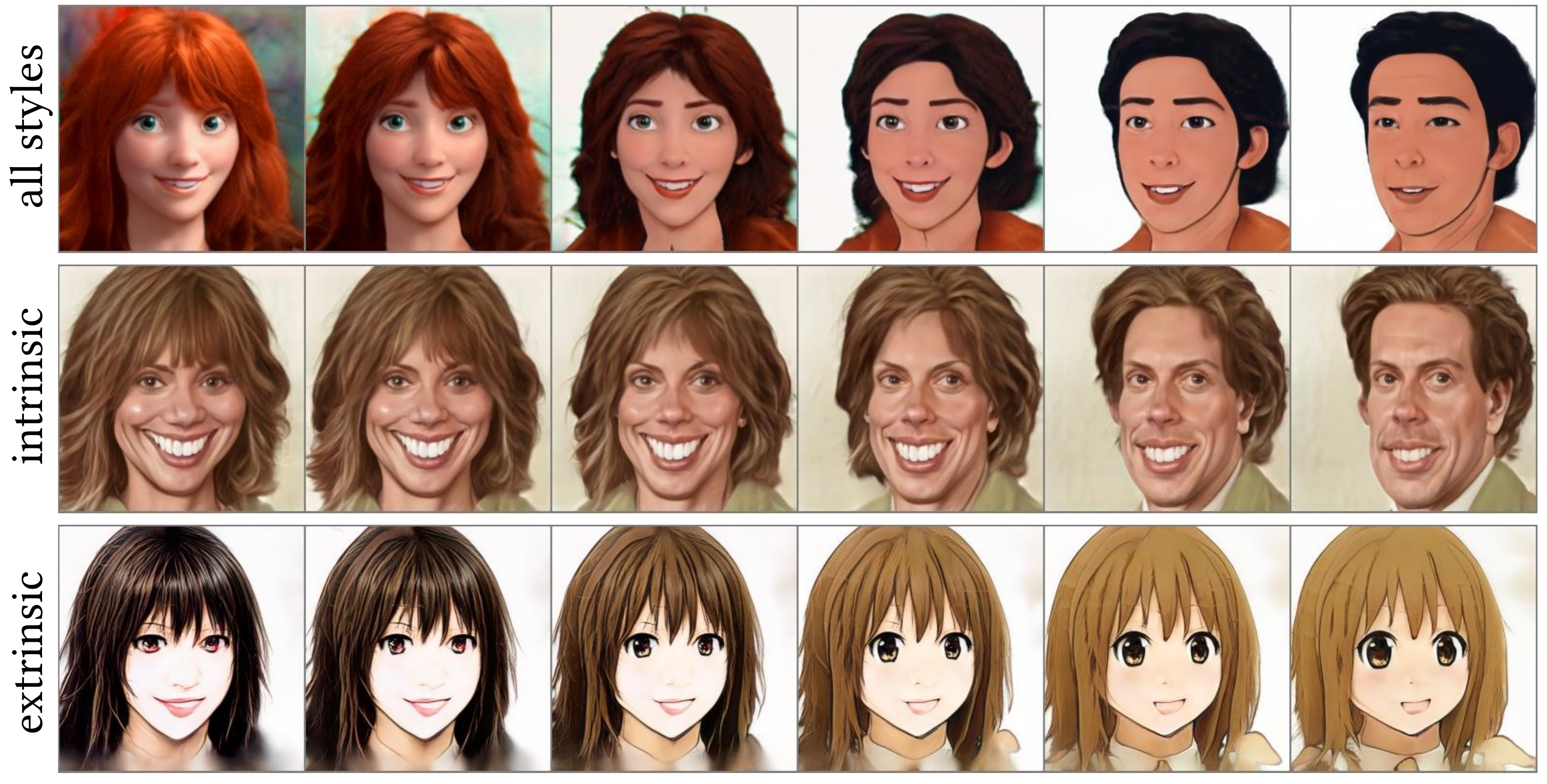}\vspace{-2mm}
\caption{Intrinsic and extrinsic style blending.}\vspace{-1em}
\label{fig:stylefusion}
\end{figure}

\noindent
\textbf{Style blending.} In Fig.~\ref{fig:stylefusion}, we fuse styles by interpolating two intrinsic and/or extrinsic style codes.
The smooth transition implies a reasonable coverage of the style manifold.

\begin{figure}[t]
\centering
\includegraphics[width=0.97\linewidth]{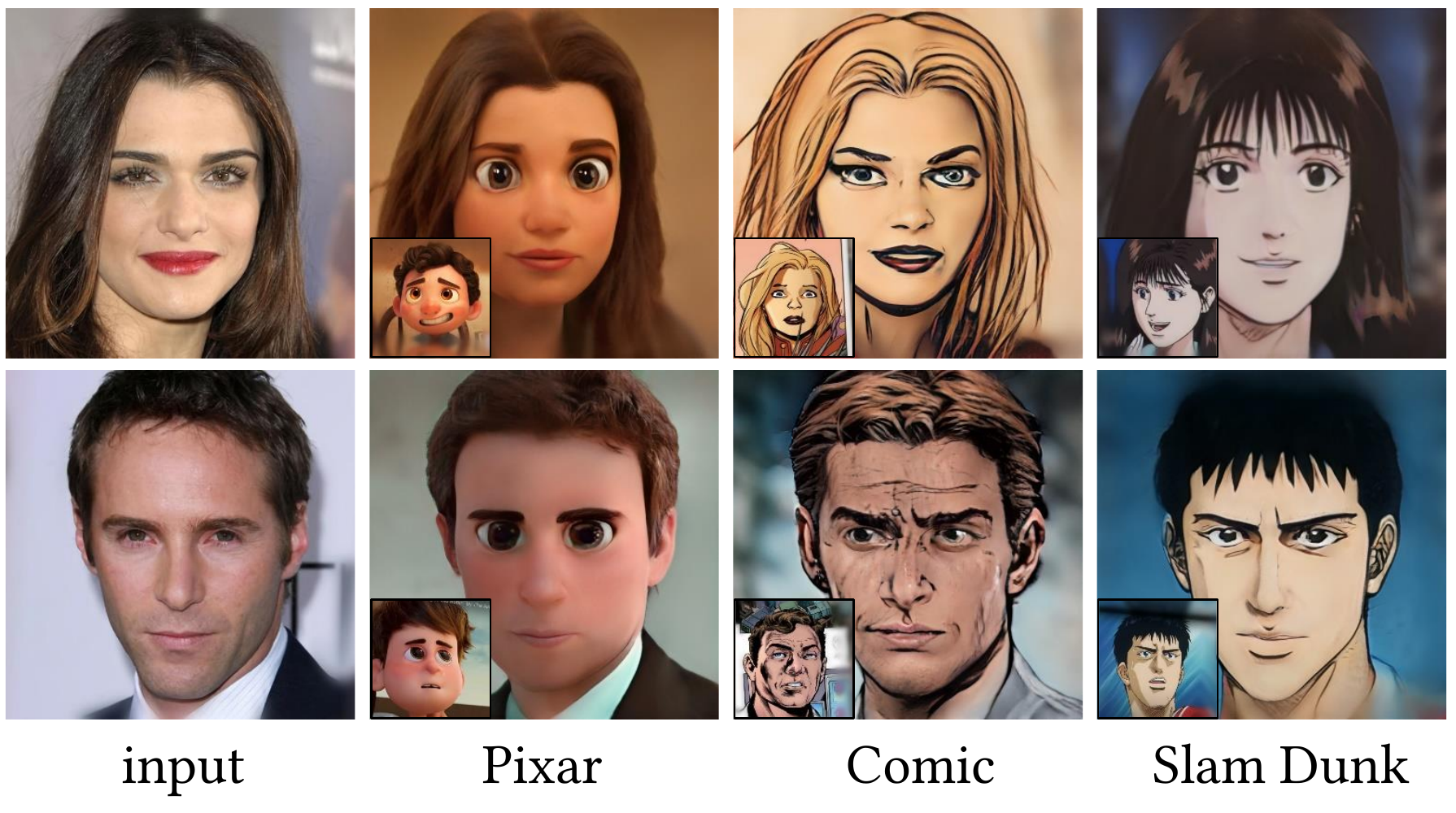}\vspace{-2.8mm}
\caption{Performance on Pixar, Comic and Slam Dunk styles.}\vspace{-0.6em}
\label{fig:more_style}
\end{figure}

\noindent
\textbf{Performance on other styles.} We further collect datasets in styles of Pixar, Comic and Slam Dunk from the Internet, with 122, 101 and 120 images, respectively. Our method achieves good performance on these styles as in Fig.~\ref{fig:more_style}.

\begin{figure}[t]
\centering
\includegraphics[width=0.97\linewidth]{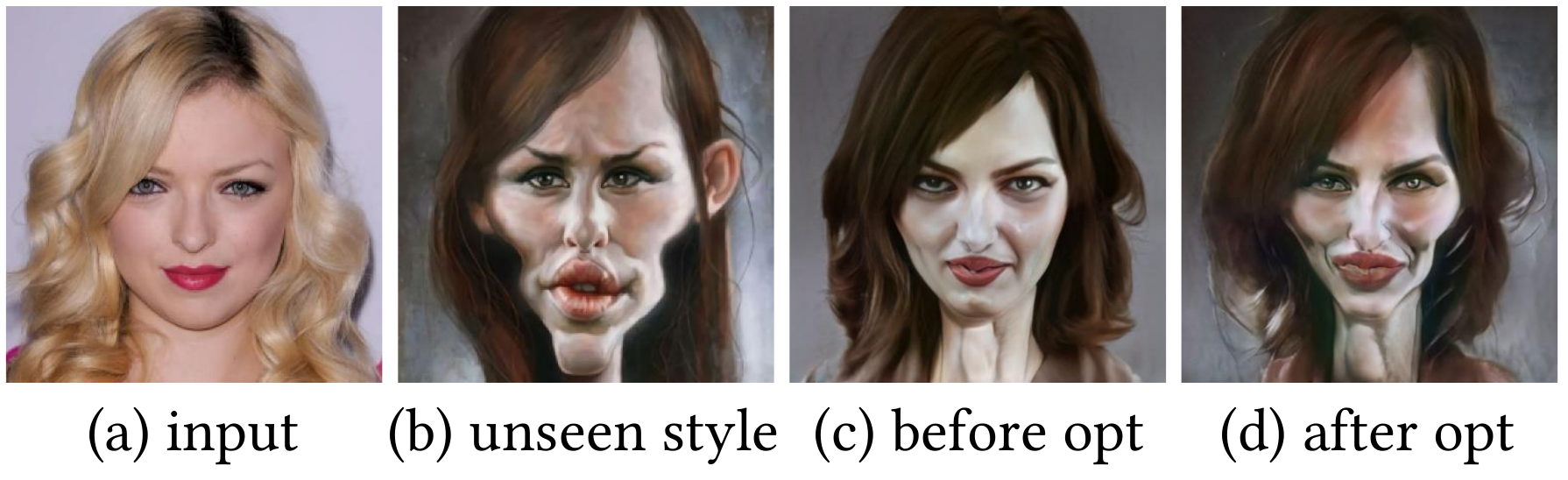}\vspace{-2.5mm}
\caption{Performance on unseen style.}\vspace{-0.8em}
\label{fig:unseen_style}
\end{figure}

\noindent
\textbf{Performance on unseen style.} Given unseen styles beyond training data, our method still transfers reasonable but less consistent styles (Fig.~\ref{fig:unseen_style}(c)).
By destylizing the unseen image to obtain a fixed intrinsic style code and optimizing the extrinsic style code as in Sec~\ref{sec:post}, better styles are learned (Fig.~\ref{fig:unseen_style}(d)). However, it introduces some artifacts. We leave robust unseen style extensions to future work.

\subsection{Limitations}
\label{sec:limitation}

\begin{figure}[t]
\centering
\includegraphics[width=0.97\linewidth]{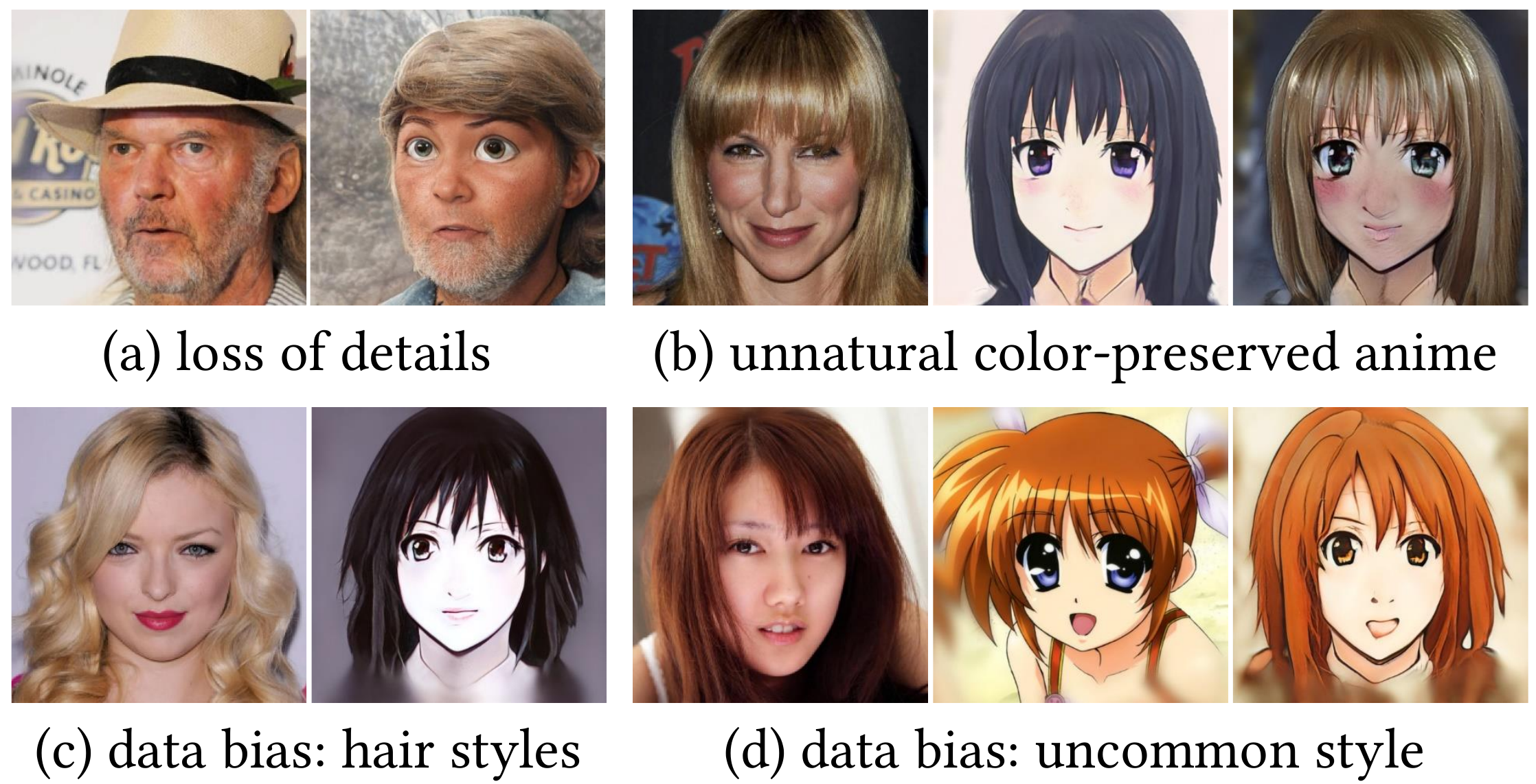}\vspace{-2mm}
\caption{Limitations of DualStyleGAN}\vspace{-1.2em}
\label{fig:limitation}
\end{figure}

In Fig.~\ref{fig:limitation}, we show three typical failure cases of DualStyleGAN.
First, while the face features are well captured, details in non-facial regions like the hat and background textures are lost in our result.
Second, anime faces often have very abstract noses. If we retain the color of the photo, the nose becomes evident but unnatural for anime style.
Third, our method still suffers from data bias problem. The anime dataset has strong bias towards straight hairs and bangs, making our method fail to handle curly hairs without bangs. Meanwhile, uncommon styles like the extremely large eyes cannot be well imitated. As a result, applying our method to tasks with severe data imbalance problem might lead to unsatisfactory results on   under-representated data.

\section{Conclusion and Future Work}\vspace{-0.5em}

In this paper, we extend StyleGAN to accept style condition from new domains while preserving its style control in the original domain. This results in an interesting application of high-resolution exemplar-based portrait style transfer with a friendly data requirement.
DualStyleGAN, with an additional style path to StyleGAN, can effectively model and modulate the intrinsic and extrinsic styles for flexible and diverse artistic portrait generation.
We show that valid transfer learning on DualStyleGAN can be achieved with especial architecture design and progressive training strategy.
We believe our idea of model extension in terms of both architecture and data can be potentially applied to other tasks such as more general image-to-image translation and knowledge distillation.
In future work, we would like to explore the recommendation of the suitable style image and its weight vector $\textbf{w}$ for the input photo for easy use and to alleviate the data bias problem via data augmentation.

\noindent
\ys{\textbf{Acknowledgments.} This study is supported under the RIE2020 Industry Alignment Fund -- Industry Collaboration Projects (IAF-ICP) Funding Initiative, as well as cash and in-kind contribution from the industry partner(s).}


{\small
\bibliographystyle{ieee_fullname}
\bibliography{egbib}

\begin{thebibliography}{10}\itemsep=-1pt

\bibitem{abdal2019image2stylegan}
Rameen Abdal, Yipeng Qin, and Peter Wonka.
\newblock Image2stylegan: How to embed images into the stylegan latent space?
\newblock In {\em Proc.~Int'l Conf.~Computer Vision}, pages 4432--4441, 2019.

\bibitem{bengio2009curriculum}
Yoshua Bengio, J{\'e}r{\^o}me Louradour, Ronan Collobert, and Jason Weston.
\newblock Curriculum learning.
\newblock In {\em Proc.~IEEE Int'l Conf.~Machine Learning}, pages 41--48, 2009.

\bibitem{danbooru2019Portraits}
Gwern Branwen, Anonymous, and Danbooru Community.
\newblock Danbooru2019 portraits: A large-scale anime head illustration
  dataset.
\newblock \url{https://www.gwern.net/Crops\#danbooru2019-portraits}, March
  2019.

\bibitem{cao2018carigans}
Kaidi Cao, Jing Liao, and Lu Yuan.
\newblock Carigans: unpaired photo-to-caricature translation.
\newblock {\em {ACM} Transactions on Graphics}, 37(6):1--14, 2018.

\bibitem{choi2020stargan}
Yunjey Choi, Youngjung Uh, Jaejun Yoo, and Jung-Woo Ha.
\newblock Stargan v2: Diverse image synthesis for multiple domains.
\newblock In {\em Proc.~IEEE Int'l Conf.~Computer Vision and Pattern
  Recognition}, pages 8188--8197, 2020.

\bibitem{chong2021gans}
Min~Jin Chong and David Forsyth.
\newblock {GANs N' Roses}: Stable, controllable, diverse image to image
  translation.
\newblock {\em arXiv preprint arXiv:2106.06561}, 2021.

\bibitem{deng2019arcface}
Jiankang Deng, Jia Guo, Niannan Xue, and Stefanos Zafeiriou.
\newblock Arcface: Additive angular margin loss for deep face recognition.
\newblock In {\em Proc.~IEEE Int'l Conf.~Computer Vision and Pattern
  Recognition}, pages 4690--4699, 2019.

\bibitem{he2016deep}
Kaiming He, Xiangyu Zhang, Shaoqing Ren, and Jian Sun.
\newblock Deep residual learning for image recognition.
\newblock In {\em Proc.~IEEE Int'l Conf.~Computer Vision and Pattern
  Recognition}, pages 770--778, 2016.

\bibitem{hoshen2019non}
Yedid Hoshen, Ke Li, and Jitendra Malik.
\newblock Non-adversarial image synthesis with generative latent nearest
  neighbors.
\newblock In {\em Proc.~IEEE Int'l Conf.~Computer Vision and Pattern
  Recognition}, pages 5811--5819, 2019.

\bibitem{huang2017adain}
Xun Huang and Serge Belongie.
\newblock Arbitrary style transfer in real-time with adaptive instance
  normalization.
\newblock In {\em Proc.~Int'l Conf.~Computer Vision}, pages 1510--1519, 2017.

\bibitem{Huo2017Variation}
Jing Huo, Yang Gao, Yinghuan Shi, and Hujun Yin.
\newblock Variation robust cross-modal metric learning for caricature
  recognition.
\newblock In {\em Proc.~Thematic Workshops of ACM Int'l Conf. Multimedia},
  pages 340--348, 2017.

\bibitem{Huo2018WebCaricature}
Jing Huo, Wenbin Li, Yinghuan Shi, Yang Gao, and Hujun Yin.
\newblock Webcaricature: a benchmark for caricature recognition.
\newblock In {\em Proc.~British Machine Vision Conference}, 2018.

\bibitem{jang2021stylecarigan}
Wonjong Jang, Gwangjin Ju, Yucheol Jung, Jiaolong Yang, Xin Tong, and Seungyong
  Lee.
\newblock {StyleCariGAN}: caricature generation via stylegan feature map
  modulation.
\newblock {\em {ACM} Transactions on Graphics}, 40(4):1--16, 2021.

\bibitem{jiang2021deceive}
Liming Jiang, Bo Dai, Wayne Wu, and Chen~Change Loy.
\newblock Deceive {D}: Adaptive pseudo augmentation for gan training with
  limited data.
\newblock In {\em Advances in Neural Information Processing Systems}, 2021.

\bibitem{Johnson2016Perceptual}
Justin Johnson, Alexandre Alahi, and Fei~Fei Li.
\newblock Perceptual losses for real-time style transfer and super-resolution.
\newblock In {\em Proc.~European Conf.~Computer Vision}, pages 694--711.
  Springer, 2016.

\bibitem{karras2018progressive}
Tero Karras, Timo Aila, Samuli Laine, and Jaakko Lehtinen.
\newblock Progressive growing of gans for improved quality, stability, and
  variation.
\newblock {\em Proc.~Int'l Conf.~Learning Representations}, 2018.

\bibitem{karras2019style}
Tero Karras, Samuli Laine, and Timo Aila.
\newblock A style-based generator architecture for generative adversarial
  networks.
\newblock In {\em Proc.~IEEE Int'l Conf.~Computer Vision and Pattern
  Recognition}, pages 4401--4410, 2019.

\bibitem{karras2020analyzing}
Tero Karras, Samuli Laine, Miika Aittala, Janne Hellsten, Jaakko Lehtinen, and
  Timo Aila.
\newblock Analyzing and improving the image quality of stylegan.
\newblock In {\em Proc.~IEEE Int'l Conf.~Computer Vision and Pattern
  Recognition}, pages 8110--8119, 2020.

\bibitem{kim2019u}
Junho Kim, Minjae Kim, Hyeonwoo Kang, and Kwang~Hee Lee.
\newblock {U-GAT-IT}: Unsupervised generative attentional networks with
  adaptive layer-instance normalization for image-to-image translation.
\newblock In {\em Proc.~Int'l Conf.~Learning Representations}, 2019.

\bibitem{kwon2021diagonal}
Gihyun Kwon and Jong~Chul Ye.
\newblock Diagonal attention and style-based gan for content-style
  disentanglement in image generation and translation.
\newblock In {\em Proc.~Int'l Conf.~Computer Vision}, 2021.

\bibitem{kwong2021unsupervised}
Sam Kwong, Jialu Huang, and Jing Liao.
\newblock Unsupervised image-to-image translation via pre-trained stylegan2
  network.
\newblock {\em {IEEE} Transactions on Multimedia}, 2021.

\bibitem{li2021anigan}
Bing Li, Yuanlue Zhu, Yitong Wang, Chia-Wen Lin, Bernard Ghanem, and Linlin
  Shen.
\newblock {AniGAN}: Style-guided generative adversarial networks for
  unsupervised anime face generation.
\newblock {\em {IEEE} Transactions on Multimedia}, 2021.

\bibitem{Li2016Combining}
Chuan Li and Michael Wand.
\newblock Combining markov random fields and convolutional neural networks for
  image synthesis.
\newblock In {\em Proc.~IEEE Int'l Conf.~Computer Vision and Pattern
  Recognition}, pages 2479--2486, 2016.

\bibitem{liao2017visual}
Jing Liao, Yuan Yao, Lu Yuan, Gang Hua, and Sing~Bing Kang.
\newblock Visual atribute transfer through deep image analogy.
\newblock {\em {ACM} Transactions on Graphics}, 36(4):120, 2017.

\bibitem{liu2015deep}
Ziwei Liu, Ping Luo, Xiaogang Wang, and Xiaoou Tang.
\newblock Deep learning face attributes in the wild.
\newblock In {\em Proc.~Int'l Conf.~Computer Vision}, pages 3730--3738, 2015.

\bibitem{mechrez2018contextual}
Roey Mechrez, Itamar Talmi, and Lihi Zelnik-Manor.
\newblock The contextual loss for image transformation with non-aligned data.
\newblock In {\em Proc.~European Conf.~Computer Vision}, pages 768--783, 2018.

\bibitem{nizan2020breaking}
Ori Nizan and Ayellet Tal.
\newblock Breaking the cycle-colleagues are all you need.
\newblock In {\em Proc.~IEEE Int'l Conf.~Computer Vision and Pattern
  Recognition}, pages 7860--7869, 2020.

\bibitem{ojha2021few}
Utkarsh Ojha, Yijun Li, Jingwan Lu, Alexei~A Efros, Yong~Jae Lee, Eli
  Shechtman, and Richard Zhang.
\newblock Few-shot image generation via cross-domain correspondence.
\newblock In {\em Proc.~IEEE Int'l Conf.~Computer Vision and Pattern
  Recognition}, pages 10743--10752, 2021.

\bibitem{pinkney2020resolution}
Justin~NM Pinkney and Doron Adler.
\newblock Resolution dependent gan interpolation for controllable image
  synthesis between domains.
\newblock {\em arXiv preprint arXiv:2010.05334}, 2020.

\bibitem{richardson2020encoding}
Elad Richardson, Yuval Alaluf, Or Patashnik, Yotam Nitzan, Yaniv Azar, Stav
  Shapiro, and Daniel Cohen-Or.
\newblock Encoding in style: a stylegan encoder for image-to-image translation.
\newblock In {\em Proc.~IEEE Int'l Conf.~Computer Vision and Pattern
  Recognition}, 2021.

\bibitem{selim2016painting}
Ahmed Selim, Mohamed Elgharib, and Linda Doyle.
\newblock Painting style transfer for head portraits using convolutional neural
  networks.
\newblock {\em {ACM} Transactions on Graphics}, 35(4):1--18, 2016.

\bibitem{shao2021spatchgan}
Xuning Shao and Weidong Zhang.
\newblock {SPatchGAN}: A statistical feature based discriminator for
  unsupervised image-to-image translation.
\newblock In {\em Proc.~Int'l Conf.~Computer Vision}, 2021.

\bibitem{shi2019warpgan}
Yichun Shi, Debayan Deb, and Anil~K Jain.
\newblock {WarpGAN}: Automatic caricature generation.
\newblock In {\em Proc.~IEEE Int'l Conf.~Computer Vision and Pattern
  Recognition}, pages 10762--10771, 2019.

\bibitem{SimonyanZ14a}
Karen Simonyan and Andrew Zisserman.
\newblock Very deep convolutional networks for large-scale image recognition.
\newblock In {\em Proc.~Int'l Conf.~Learning Representations}, 2015.

\bibitem{song2021agilegan}
Guoxian Song, Linjie Luo, Jing Liu, Wan-Chun Ma, Chunpong Lai, Chuanxia Zheng,
  and Tat-Jen Cham.
\newblock Agilegan: stylizing portraits by inversion-consistent transfer
  learning.
\newblock {\em {ACM} Transactions on Graphics}, 40(4):1--13, 2021.

\bibitem{tewari2020pie}
Ayush Tewari, Mohamed Elgharib, Florian Bernard, Hans-Peter Seidel, Patrick
  P{\'e}rez, Michael Zollh{\"o}fer, and Christian Theobalt.
\newblock {PIE}: Portrait image embedding for semantic control.
\newblock {\em {ACM} Transactions on Graphics}, 39(6):1--14, 2020.

\bibitem{wu2021stylealign}
Zongze Wu, Yotam Nitzan, Eli Shechtman, and Dani Lischinski.
\newblock {StyleAlign}: Analysis and applications of aligned stylegan models.
\newblock {\em arXiv preprint arXiv:2110.11323}, 2021.

\bibitem{xie2021unaligned}
Shaoan Xie, Mingming Gong, Yanwu Xu, and Kun Zhang.
\newblock Unaligned image-to-image translation by learning to reweight.
\newblock In {\em Proc.~Int'l Conf.~Computer Vision}, pages 14174--14184, 2021.

\bibitem{zhao2020unpaired}
Yihao Zhao, Ruihai Wu, and Hao Dong.
\newblock Unpaired image-to-image translation using adversarial consistency
  loss.
\newblock In {\em Proc.~European Conf.~Computer Vision}, pages 800--815.
  Springer, 2020.

\bibitem{Zhu2017Unpaired}
Jun~Yan Zhu, Taesung Park, Phillip Isola, and Alexei~A. Efros.
\newblock Unpaired image-to-image translation using cycle-consistent
  adversarial networks.
\newblock In {\em Proc.~Int'l Conf.~Computer Vision}, pages 2242--2251, 2017.

\end{thebibliography}
}

\clearpage

\appendix

\section{Appendix: Implementation Details}

\subsection{Dataset}

Cartoon dataset~\cite{pinkney2020resolution} has 317 images, which is provided at~\url{https://github.com/justinpinkney/toonify}.
We use 199 images from WebCaricature~\cite{Huo2017Variation,Huo2018WebCaricature} to build the Caricature dataset.
We use 174 images from Danbooru Portraits~\cite{danbooru2019Portraits} to build the Anime dataset.
We follow the FFHQ~\cite{karras2019style} to align caricature and anime faces based on facial landmarks.
The landmarks of caricature faces are included in WebCaricature. The landmarks of anime faces are manually labelled by us.

\subsection{Network Architecture}

There are 7 ModRes blocks in the extrinsic style path, corresponding to one $4\times4$, two $8\times8$, two $16\times16$ and two $32\times32$ convolution layers in StyleGAN.
Each ModRes has a ResBlock and an AdaIN block. Each ResBlock has two convolution layers with $3\times3$  kernels.
The dimension of the style code of AdaIN is the same as the intermediate style code $\mathbf{w}$ of StyleGAN, \ie, $512$.
The structure transfer block $T_s$ in the extrinsic style path is made up of two linear layers.
There are 11 color transform block $T_c$ in the extrinsic style path, corresponding to two $64\times64$, two $128\times128$, two $256\times256$,  two $512\times512$, two $1024\times1024$ convolution layers and one $1024\times1024$ ToRGB layer in StyleGAN.
$T_c$ is made up of a linear layer.

\subsection{Network Training}

\noindent
\textbf{Destylization -- Stage II} uses one NVIDIA Tesla V100 GPU, and optimizes the latent code for 300 iterations, which processes about 45 images per hour.

\noindent
\textbf{Progressive fine-tuning -- Stage II} uses one NVIDIA Tesla V100 GPU and a batch size of $4$ per GPU with $\lambda_{\text{adv}}=0.1,\lambda_{\text{perc}}=0.5$.
To calculate $\lambda_{\text{perc}}$, we use the conv1\_1, conv2\_1 and conv3\_1 layers of the VGG19~\cite{SimonyanZ14a} with equal weights of $1$.
We train on $l=7$, $6$, $5$ for 300, 300, 3000 iterations, respectively, Stage II takes about 1.8 hour. Note that once trained, the pre-trained model can be applied to any style dataset.

\noindent
\textbf{Progressive fine-tuning -- Stage III} uses 8 NVIDIA Tesla V100 GPUs and a batch size of $4$ per GPU with  $\lambda_{\text{adv}}=1,\lambda_{\text{perc}}=1,\lambda_{\text{CX}}=0.25,\lambda_{\text{FM}}=0.25$.
To calculate $\lambda_{\text{perc}}$, we use the conv2\_1 and conv3\_1 layers of the VGG19~\cite{SimonyanZ14a} weighted by $0.5$ and $1$, respectively
We set $(\lambda_{\text{ID}}, \lambda_{\text{reg}})$ to $(1,0.015)$, $(4,0.005)$, $(1,0.02)$ and trains for 1400, 1000, 2100 iterations on cartoon, caricature and anime, respectively. Training takes about 0.75 hour on average.

\noindent
\textbf{Post-processing}. Latent optimization and training sampling network use one GPU.
Post-processing optimizes the latent for 100 iterations, taking about 0.46 hour per 100 artistic portrait images.
We adopt the Adam optimizer.
The first 7 rows of $\mathbf{z}^+_e$ use a small learning rate of $0.005$, $0.005$, $0.0005$ to prevent violent structure changes for cartoon, caricature and anime, respectively.
The last 11 rows of $\mathbf{z}^+_e$  use a learning rate of $0.1$, $0.01$, $0.005$ for color refinement for cartoon, caricature and anime, respectively. Training two sampling networks takes about 0.13 hour.

\section{Appendix: Experimental Details of Simulating Fine-Tuning Behavior}

\begin{figure*}[htbp]
\centering
\includegraphics[width=0.8\linewidth]{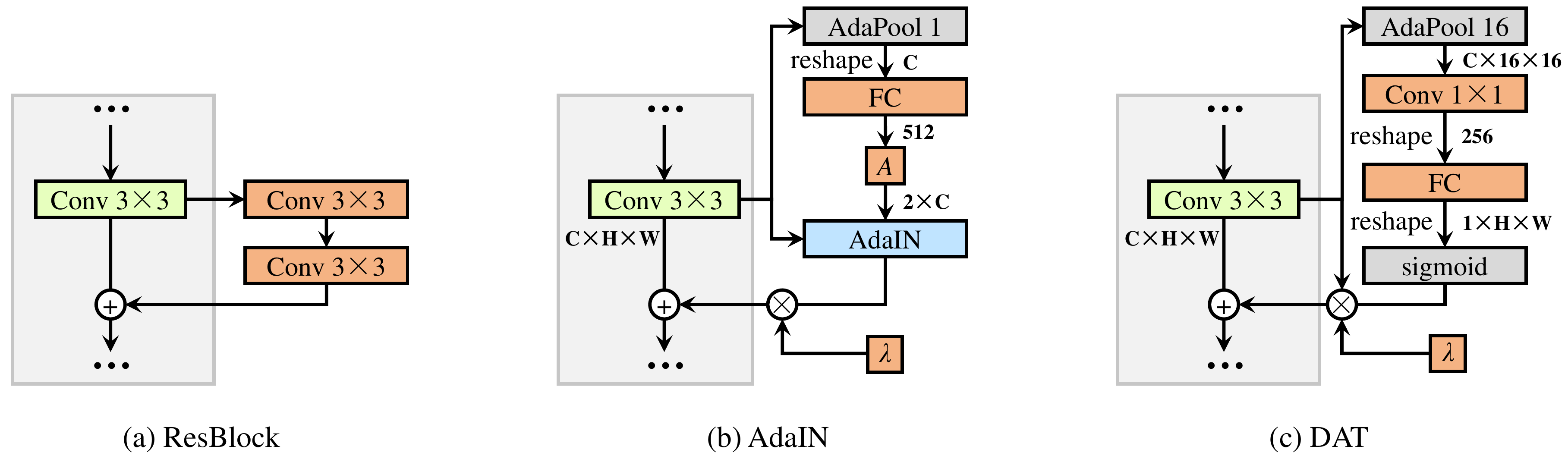}\vspace{-2mm}
\caption{Network architectures used to simulate fine-tuning behavior.}\vspace{-2mm}
\label{fig:resblock}
\end{figure*}

As analyzed in Sec.~\ref{sec:network} of the paper, StyleGANs before and after fine-tuning have shared latent spaces~\cite{kwong2021unsupervised} and closely-related convolution features. Therefore, the difference of these convolution features is also closely-related to the original convolution features.
It is possible to keep all other submodules fixed but only learn changes over the convolution features to simulate the changes of the convolution weight matrices in fine-tuning, which naturally corresponding to a residual path.
To simulate the unconditional fine-tuning over StyleGAN, we follow three principles:
\begin{itemize}[itemsep=1.5pt,topsep=1pt,parsep=0pt]
  \item \textbf{Principle I}: The residual path has few impact on the feature at the beginning of fine-tuning in order to make the pre-trained generative space unaltered.
  \item \textbf{Principle II}: The residual path is conditioned by the input convolution features from StyleGAN, considering the residual features should be closely-related to the original convolution features.
  \item \textbf{Principle III}: The residual path is not conditioned by the external style images, since in this experiment we would like to only simulate the unconditional fine-tuning over StyleGAN rather than learning a style transfer task.
\end{itemize}

Based on the principles, we compare three modules for building the residual path: channel-wise AdaIN~\cite{huang2017adain},
spatial-wise Diagonal Attention (DAT)~\cite{kwon2021diagonal} and element-wise ResBlock~\cite{he2016deep}, with the architectures as in Fig.~\ref{fig:resblock}:
\begin{itemize}[itemsep=1.5pt,topsep=1pt,parsep=0pt]
  \item \textbf{ResBlock}: Standard residual blocks with two $3\times3$ convolution layers. The convolution weight matrices are set to values close to 0 to meet Principle I.
  \item \textbf{AdaIN}: To meet Principles II and III, we extract the channel-wise style code from the input convolution feature by global average pooling and a linear layer rather than using the external style code. The style code then goes through an affine transform block $A$ for adaptive instance normalization. To meet Principle I, the modulated feature is multiplied with a learnable weight $\lambda$, which is set to a small value of $0.01$ initially, before added to the input feature.
    \item \textbf{DAT}: To meet Principles II and III, we extract the spatial-wise style code from the  input convolution feature  via an adaptive pooling layer and an $1\times1$ convolution layer rather than using the external style code. The one-channel style code goes through a linear layer and a sigmoid layer to obtain the attention map as the original DAT. The attention map is multiplied with the input feature as well as the learnable weight $\lambda$, which is set to a small value of $0.01$ initially, before added to the input feature.
\end{itemize}

\begin{figure}[htbp]
\centering
\includegraphics[width=0.99\linewidth]{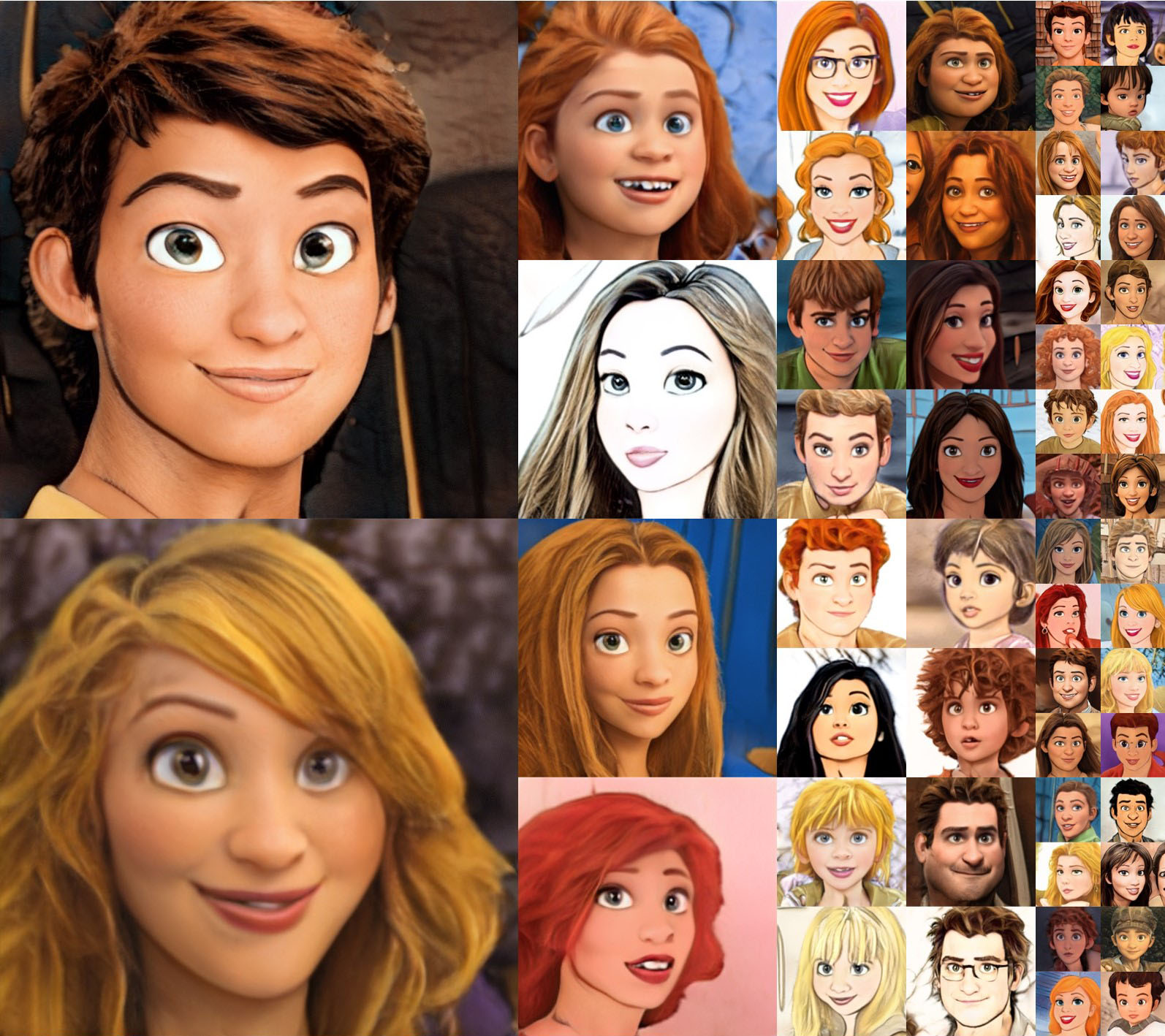}
\caption{Random cartoon faces generated by DualStyleGAN.}\vspace{3mm}
\label{fig:generation1}
\includegraphics[width=0.99\linewidth]{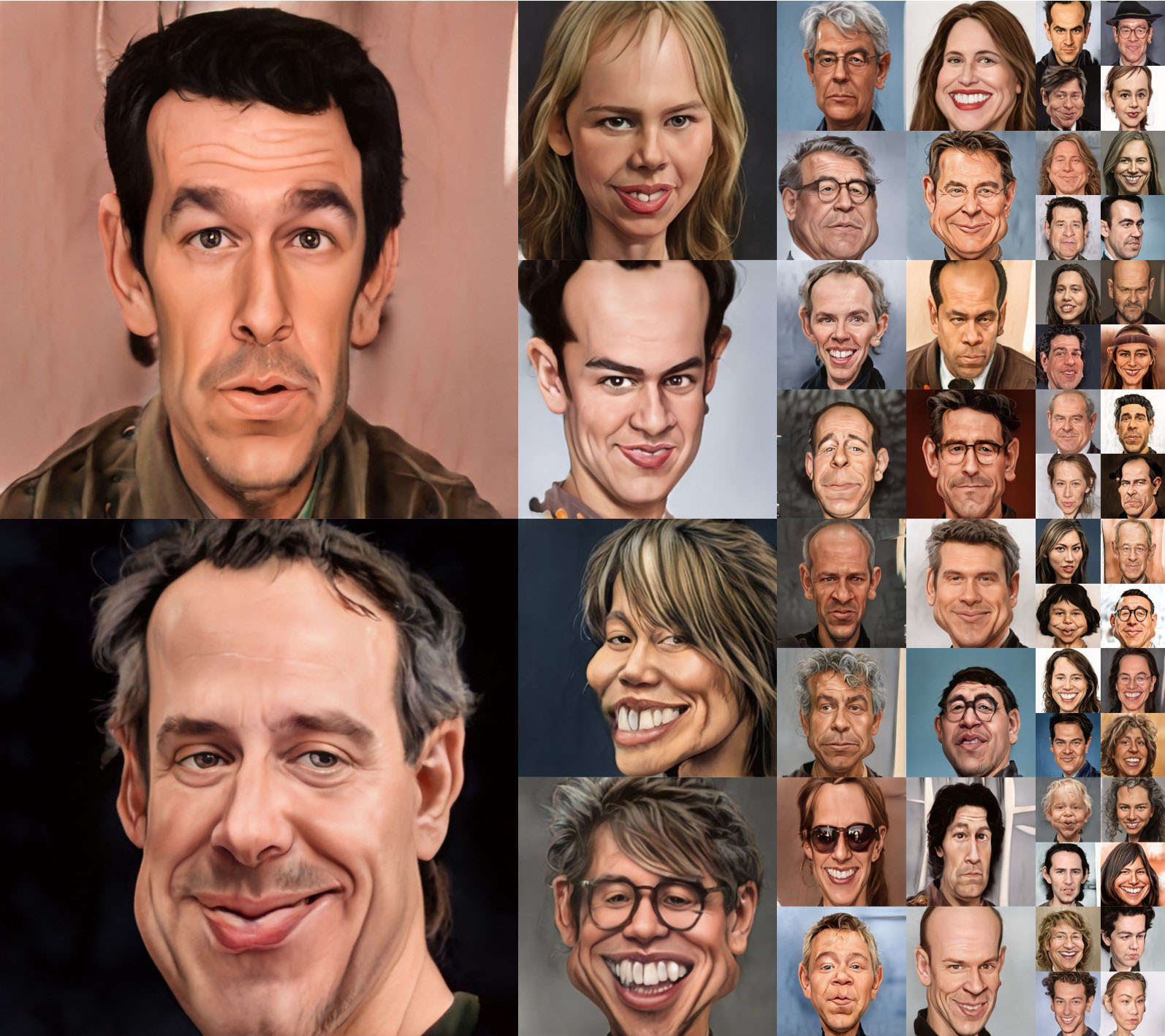}
\caption{Random caricature faces generated by DualStyleGAN.}
\label{fig:generation2}
\end{figure}

In the experiment of Sec.~\ref{sec:network} of the paper, we finetune StyleGAN for 900 iterations on Cartoon dataset. The resulting StyleGAN serves as a ground truth model.
Then, we keep the original StyleGAN fixed, and only train three kinds of residual paths for 900 iterations on Cartoon dataset.
Finally, we compare the performance of the three model against the ground truth model in cartoon face generation from the same random latent code in Fig.~\ref{fig:path}, and find that ResBlocks achieve the most similar results to those by the ground truth model. Therefore, we choose residual blocks to build the residual path in the paper.

\section{Appendix: Supplementary Experiment}

\subsection{Arbitrary Artistic Portrait Generation}

\begin{figure}[htbp]
\centering
\includegraphics[width=0.99\linewidth]{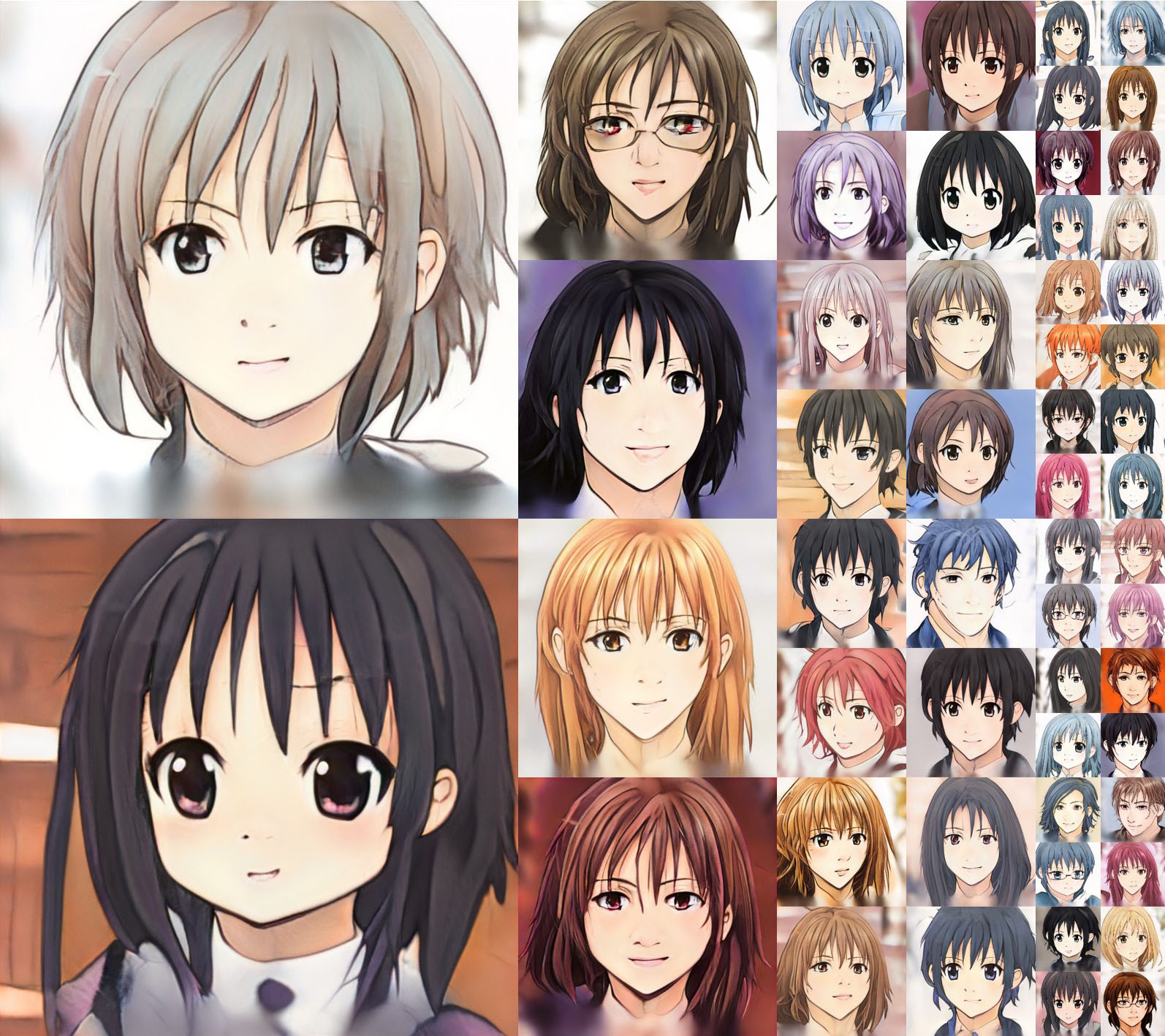}\vspace{-2mm}
\caption{Random anime faces generated by DualStyleGAN}\vspace{-2mm}
\label{fig:generation3}
\end{figure}

Figs.~\ref{fig:generation1}--\ref{fig:generation3} show artistic portraits generated by DualStyleGAN from random intrinsic and extrinsic style codes.

\subsection{Comparison with StyleCariGAN}

The officially release model of StyleCariGAN~\cite{jang2021stylecarigan} is trained on 6K images to produce $256\times256$ images. The reason might be it is difficult to train cycle translation with limited high-resolution images.
As for style control, StyleCariGAN uses style mixing for exemplar-based style control, which is limited to color control like UI2I-style~\cite{kwong2021unsupervised}.
The cycle translation of StyleCariGAN is not conditioned on example. Thus, it learns an overall structure transfer rather than exemplar-based structure transfer. In~\cite{jang2021stylecarigan}, StyleCariGAN shows exemplar-based structure transfer results by mixing the structure style at the low-resolution layers, which is however at the cost of lost identity and artifacts.

\begin{figure}[htbp]
\centering
\includegraphics[width=0.99\linewidth]{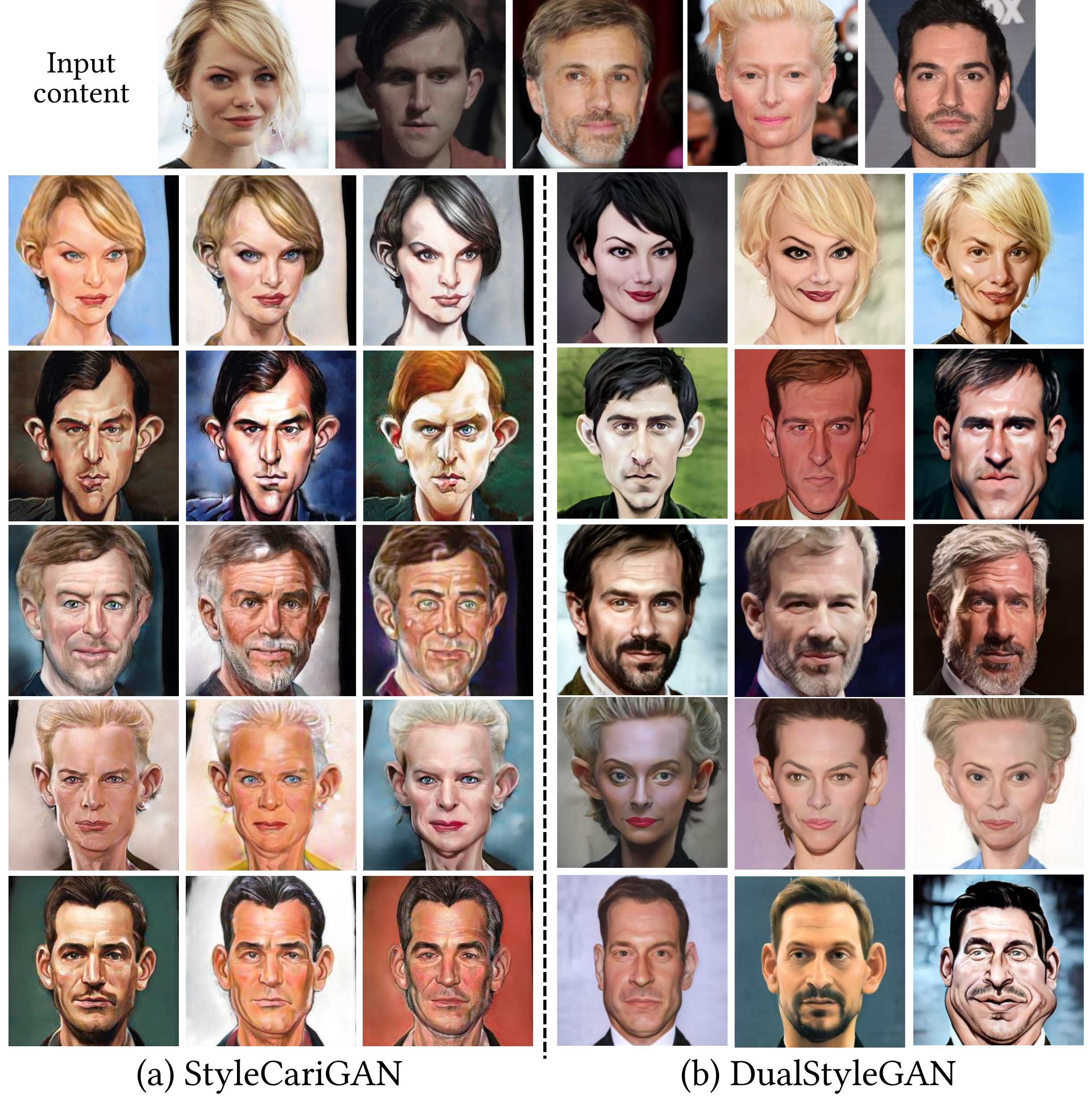}
\caption{Random anime faces generated by DualStyleGAN}\vspace{-4mm}
\label{fig:stylecarigan2}
\end{figure}

In Fig.~\ref{fig:stylecarigan}, we compare with StyleCariGAN, \ys{where the performance of StyleCariGAN degrades when using style codes from example images. The reason might be that the style code is optimized in $\mathcal{W}+$ space as in~\cite{jang2021stylecarigan} and is out-of-distribution,
while the style codes in official style palette are directly drawn from the standard latent distribution.}
Figure~\ref{fig:stylecarigan2} presents more visual comparison results on the content images from \url{https://github.com/wonjongg/StyleCariGAN}. For StyleCariGAN, we only use style codes from official style palette to prevent degradation. For DualStyleGAN, we use the style codes from our Caricature dataset.
It can be seen that both methods render diverse and plausible textures and colors. Our method surpasses StyleCariGAN in transferring diverse structure styles.

\begin{figure}[htbp]
\centering
\includegraphics[width=0.97\linewidth]{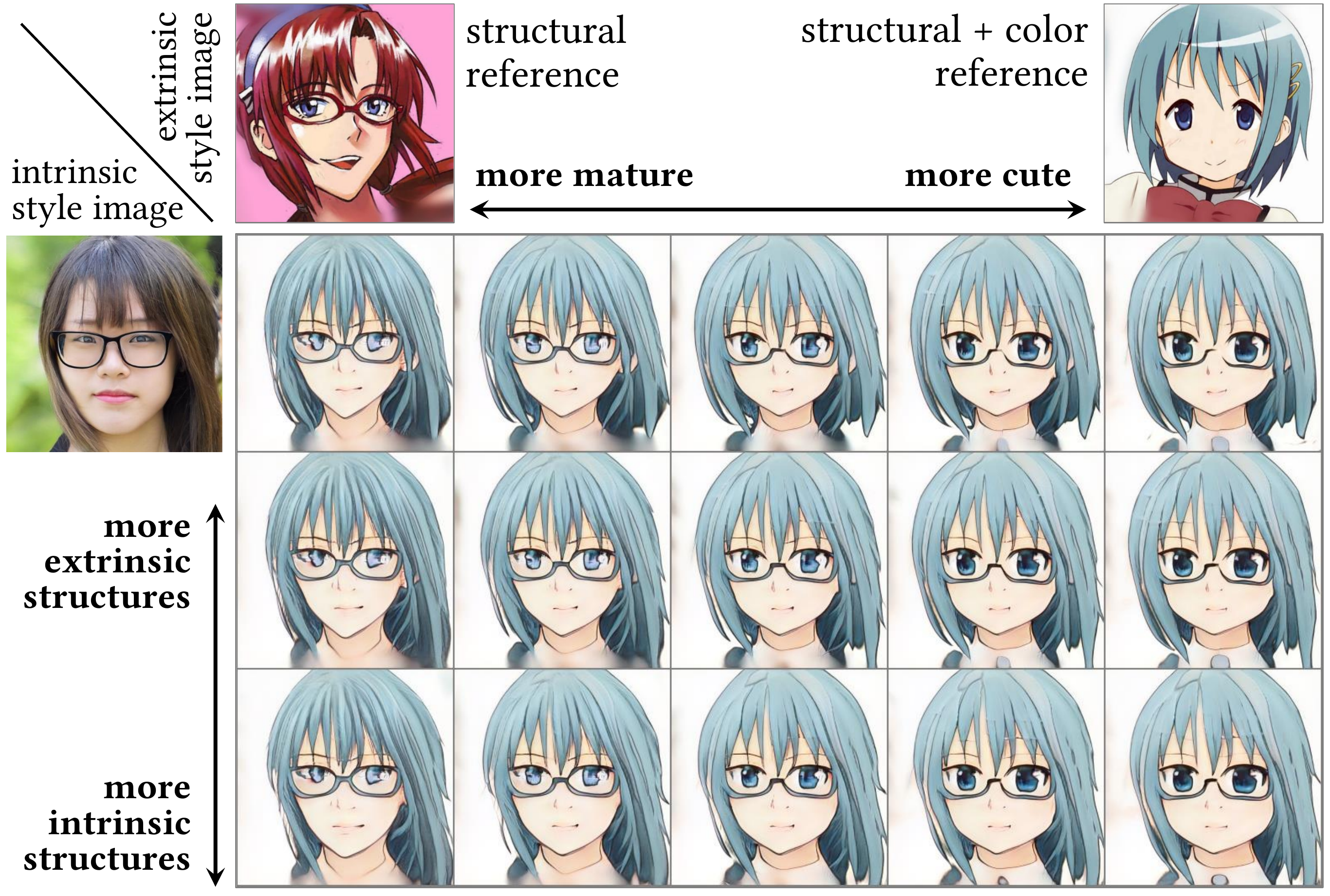}\vspace{-2mm}
\caption{Flexible style manipulation.}\vspace{-2mm}
\label{fig:blending}
\end{figure}

\begin{table} [h]
\caption{User preference ratio of state-of-the-art methods.}\vspace{-2mm}
\label{tb:1}
\centering
\scriptsize
\begin{tabular}{c|c|cccc}
\toprule
Style & ID & DualStyleGAN & UI2I-style & StarGANv2 & GNR \\
\midrule
\multirowcell{10}{Cartoon} & 1 & \textbf{0.96}  & 0.04  & 0.00  & 0.00 \\
 & 2 & \textbf{1.00}  & 0.00  & 0.00  & 0.00 \\
 & 3 & \textbf{0.93}  & 0.00  & 0.07  & 0.00 \\
 & 4 & \textbf{0.96}  & 0.00  & 0.04  & 0.00 \\
 & 5 & \textbf{1.00}  & 0.00  & 0.00  & 0.00 \\
 & 6 & \textbf{1.00}  & 0.00  & 0.00  & 0.00 \\
 & 7 & \textbf{0.89}  & 0.11  & 0.00  & 0.00 \\
 & 8 & \textbf{0.96}  & 0.04  & 0.00  & 0.00 \\
 & 9 & \textbf{0.85}  & 0.04  & 0.04  & 0.07 \\
 & 10 & \textbf{0.74}  & 0.26  & 0.00  & 0.00 \\
 \midrule
\multirowcell{10}{Caricature} & 1 & \textbf{0.78}  & 0.19  & 0.00  & 0.04 \\
 & 2 & \textbf{0.93}  & 0.00  & 0.00  & 0.07 \\
 & 3 & \textbf{0.74}  & 0.04  & 0.00  & 0.22 \\
 & 4 & \textbf{0.52}  & 0.48  & 0.00  & 0.00 \\
 & 5 & \textbf{0.96}  & 0.00  & 0.00  & 0.04 \\
 & 6 & \textbf{0.78}  & 0.19  & 0.00  & 0.04 \\
 & 7 & \textbf{0.89}  & 0.07  & 0.00  & 0.04 \\
 & 8 & \textbf{0.63}  & 0.33  & 0.00  & 0.04 \\
 & 9 & \textbf{0.81}  & 0.11  & 0.00  & 0.07 \\
 & 10 & \textbf{0.81}  & 0.07  & 0.04  & 0.07 \\
 \midrule
\multirowcell{10}{Anime} & 1 & \textbf{0.70}  & 0.07  & 0.22  & 0.00 \\
 & 2 & \textbf{0.96}  & 0.04  & 0.00  & 0.00 \\
 & 3 & \textbf{0.67}  & 0.26  & 0.00  & 0.07 \\
 & 4 & \textbf{0.33}  & 0.30  & 0.11  & 0.26 \\
 & 5 & \textbf{0.89}  & 0.07  & 0.00  & 0.04 \\
 & 6 & \textbf{0.74}  & 0.22  & 0.04  & 0.00 \\
 & 7 & \textbf{0.81}  & 0.19  & 0.00  & 0.00 \\
 & 8 & \textbf{0.85}  & 0.15  & 0.00  & 0.00 \\
 & 9 & \textbf{0.85}  & 0.15  & 0.00  & 0.00 \\
 & 10 & \textbf{1.00}  & 0.00  & 0.00  & 0.00 \\
 \midrule
 Average &  & \textbf{0.83}  & 0.11  & 0.02  & 0.04 \\
\bottomrule
\end{tabular}
\end{table}

\subsection{Style Manipulation}

In Fig.~\ref{fig:blending}, we show an example of comprehensive style manipulation, where
users are free to control whether the structure style is more \textit{mo\'{e}} (cuteness in Japanese), more mature or more in line with the hair and face shape in the photo, while keeping the cute color style.

\begin{figure*}[htbp]
\centering
\includegraphics[width=0.99\linewidth]{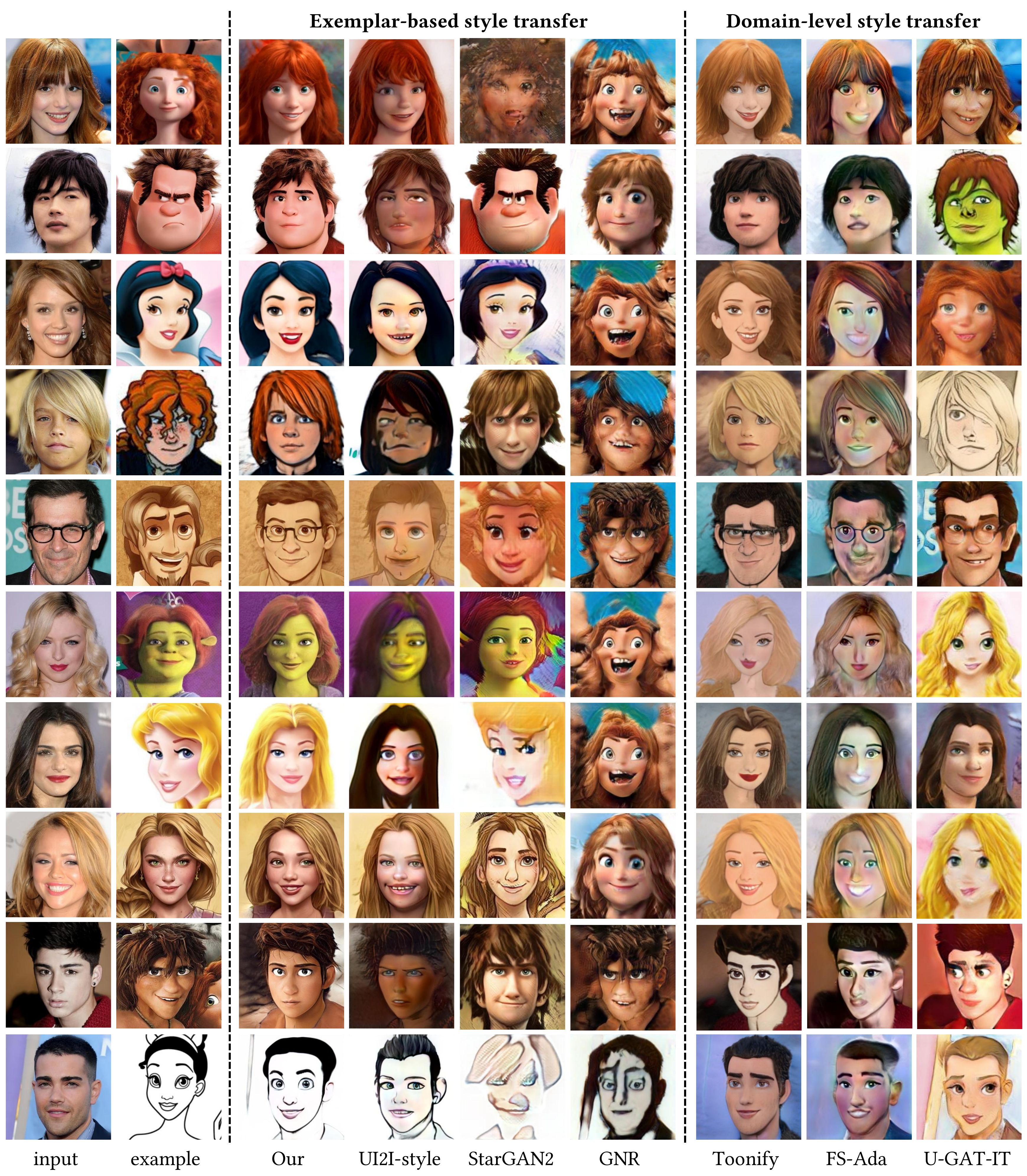}
\caption{Visual comparison on cartoon face style transfer.}
\label{fig:comparison1}
\end{figure*}

\subsection{Comparison with State-of-the-Art Methods}

\begin{figure*}[htbp]
\centering
\includegraphics[width=0.99\linewidth]{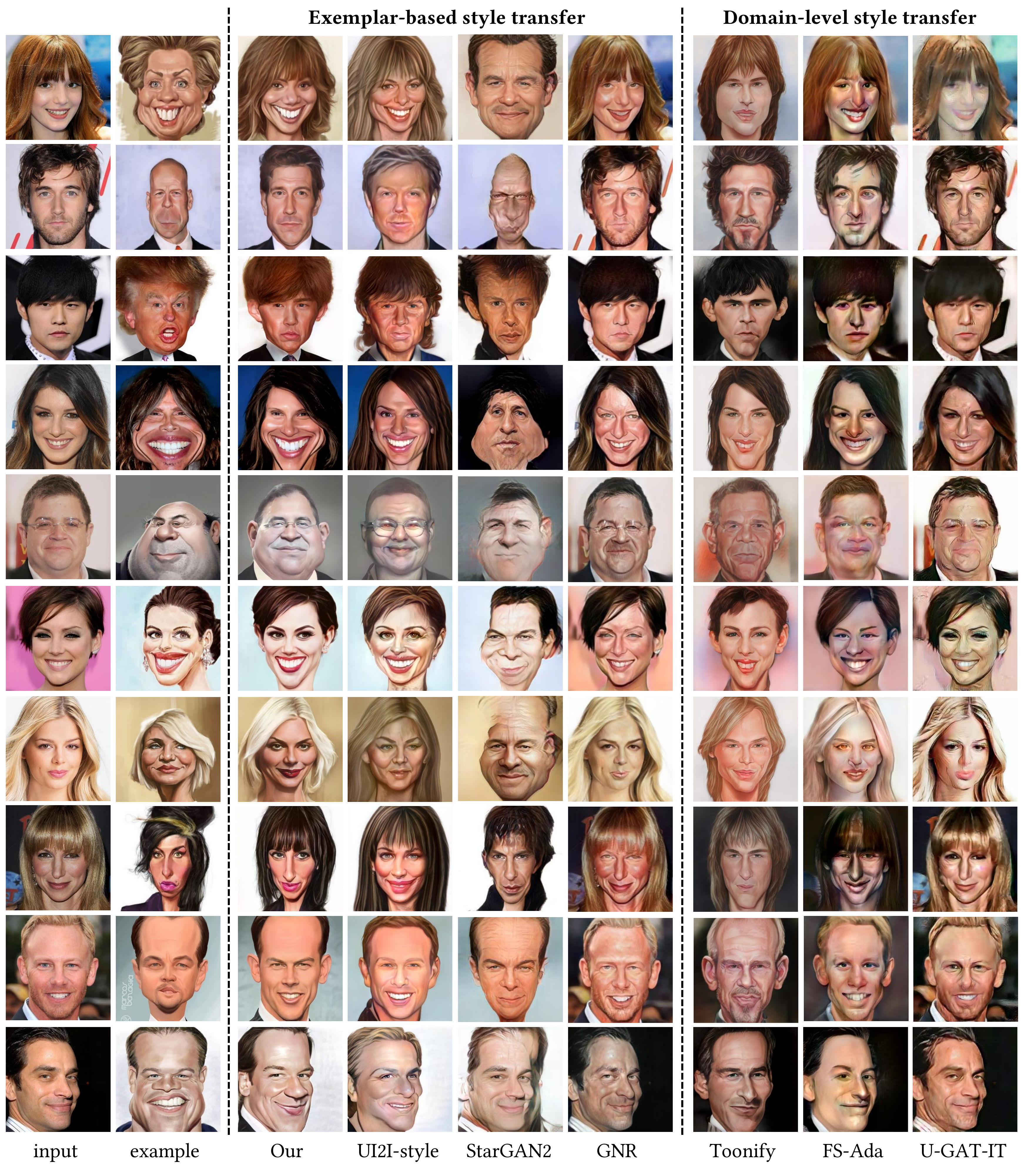}
\caption{Visual comparison on caricature face style transfer.}
\label{fig:comparison2}
\end{figure*}

\begin{figure*}[htbp]
\centering
\includegraphics[width=0.99\linewidth]{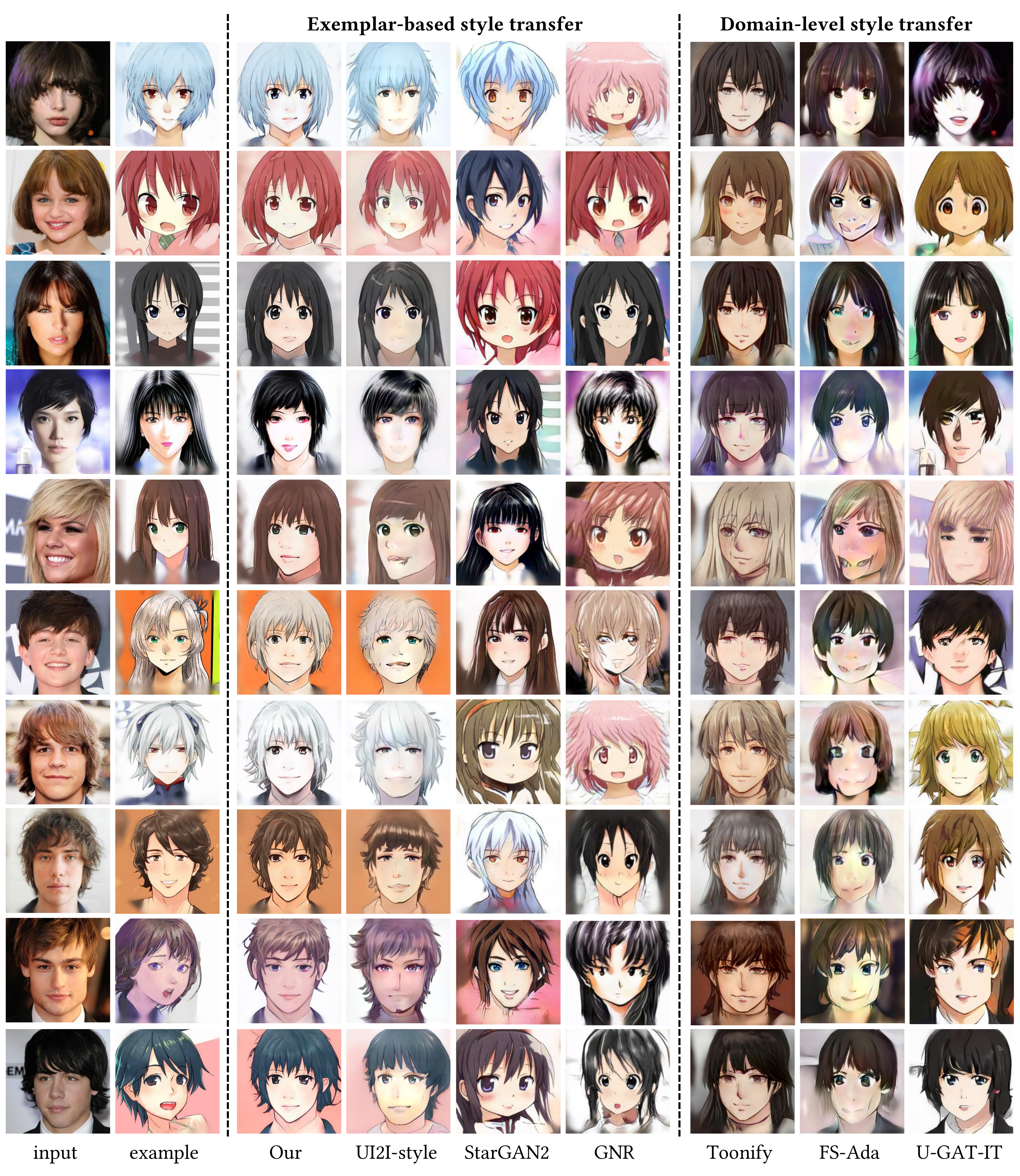}
\caption{Visual comparison on anime face style transfer.}
\label{fig:comparison3}
\end{figure*}

\noindent
\textbf{User study.} In the user study in Sec.~\ref{sec:comparison},
users are asked to select the best style transfer result in terms of both content preservation and style consistency with the reference style images.
A total of 27 subjects participate in this study to select the best ones from the results of four methods.
A total of 810 selections on 30 groups of results (Figs.~\ref{fig:comparison1}-\ref{fig:comparison3}) are tallied. Table~\ref{tb:1} demonstrates the average user scores of each group and the whole 30 groups, where the proposed method receives notable preference.

\noindent
\textbf{Visual comparison.}
In addition to the examples shown in Sec.~\ref{sec:comparison}, we show full visual comparison results used for user study in Figs.~\ref{fig:comparison1}-\ref{fig:comparison3}.

\end{document}